\documentclass[lettersize, journal]{IEEEtran}

\usepackage{graphicx}
\usepackage{subfigure}
\usepackage{multirow}
\usepackage{amssymb}
\usepackage{amsmath}
\usepackage{xcolor}
\usepackage{color}
\usepackage{siunitx}
\usepackage{bbm}
\usepackage[ruled]{algorithm2e}
\usepackage{cite}
\usepackage{diagbox}
\usepackage{tabularx}
\usepackage{hyperref} 

\usepackage{prettyref}
\newrefformat{fig}{Figure~\ref{#1}}
\newrefformat{sec}{Section~\ref{#1}}
\newrefformat{tab}{Table~\ref{#1}}
\newrefformat{algo}{Algorithm~\ref{#1}}
\newrefformat{eq}{Equation~\ref{#1}}

\graphicspath{
    {./fig/}
    }

\begin{document}

\title{PCKRF: Point Cloud Completion and Keypoint Refinement with Fusion Data for 6D Pose Estimation}

\author{ Yiheng Han$^{1}$$^{*}$, Irvin Haozhe Zhan$^{2}$$^{*}$, Long Zeng$^{3}$, Yu-Ping Wang$^{4}$, Ran Yi$^{5}$, Minjing Yu$^{6}$, Matthieu Gaetan Lin$^{2}$, Jenny Sheng$^{2}$ and Yong-Jin Liu$^{\dag}$
\thanks{This work was partially supported by Beijing Natural Science Foundation (L222008) and Natural Science Foundation of Guangdong Province (2022A1515011234).}
\thanks{$^{1}$ Yiheng Han is with the Faculty of Information Technology, Beijing University of Technology, Beijing, China.
        {\tt\small hanyiheng@bjut.edu.cn}}
\thanks{$^{2}$I.H. Zhan, M.G. Lin, J. Sheng and Y-J Liu are with BNRist, MOE-Key Laboratory of Pervasive Computing, Department of Computer Science and Technology, Tsinghua University, Beijing, China.
        {\{\tt\small zhanhz20@mails.  yh-lin21@mails. cqq22@mails. and liuyongjin@\}tsinghua.edu.cn}}
\thanks{$^{3}$Long Zeng is with the Department of Advanced Manufacturing, Shenzhen International Graduate School, Tsinghua University, Shenzhen, China.
        {\tt\small zenglong@sz.tsinghua.edu.cn}}
\thanks{$^{4}$Y-P Wang is with the Beijing Institute of Technology, Beijing, China.
        {\tt\small wyp\_cs@bit.edu.cn }}
\thanks{$^{5}$Ran Yi is with the Shanghai Jiao Tong University, Shanghai, China.
        {\tt\small ranyi@sjtu.edu.cn}}
\thanks{$^{6}$Minjing Yu is with the College of Intelligence and Computing, Tianjin University, Tianjin, China.
        {\tt\small minjingyu@tju.edu.cn}}
\thanks{$^{*}$Joint first authors  $^{\dag}$Corresponding author}
}



\maketitle


\begin{abstract}

\color{black}Some robust point cloud registration approaches with controllable pose refinement magnitude, such as ICP and its variants, are commonly used to improve 6D pose estimation accuracy. However, the effectiveness of these methods gradually diminishes with the advancement of deep learning techniques and the enhancement of initial pose accuracy, primarily due to their lack of specific design for pose refinement. In this paper, we propose Point Cloud Completion and Keypoint Refinement with Fusion Data (PCKRF), a new pose refinement pipeline for 6D pose estimation. The pipeline consists of two steps. First, it completes the input point clouds via a novel pose-sensitive point completion network. The network uses both local and global features with pose information during point completion. Then, it registers the completed object point cloud with the corresponding target point cloud by our proposed Color supported Iterative KeyPoint (CIKP) method. The CIKP method introduces color information into registration and registers a point cloud around each keypoint to increase stability. The PCKRF pipeline can be integrated with existing popular 6D pose estimation methods, such as the full flow bidirectional fusion network, to further improve their pose estimation accuracy. Experiments demonstrate that our method exhibits superior stability compared to existing approaches when optimizing initial poses with relatively high precision. Notably, the results indicate that our method effectively complements most existing pose estimation techniques, leading to improved performance in most cases. Furthermore, our method achieves promising results even in challenging scenarios involving textureless and symmetrical objects. Our source code is available at https://github.com/zhanhz/KRF. \color{black}
\end{abstract}


\renewcommand{\arraystretch}{1.2}

\section{Introduction}
\IEEEPARstart{6}{D} object pose estimation is an essential component in various applications, including robotic manipulation\cite{zhu2014single,tremblay2018deep}, augmented reality\cite{marchand2015pose}, and autonomous driving\cite{chen2017multi, Xu_2018_CVPR}. It has received extensive attention and has led to many research works over the past decade. Nonetheless, the task presents considerable challenges due to sensor noise, occlusion between objects, varying lighting conditions, and symmetries of objects.

 \begin{figure}[t]
  \centering
  {\includegraphics[width=0.9\linewidth]{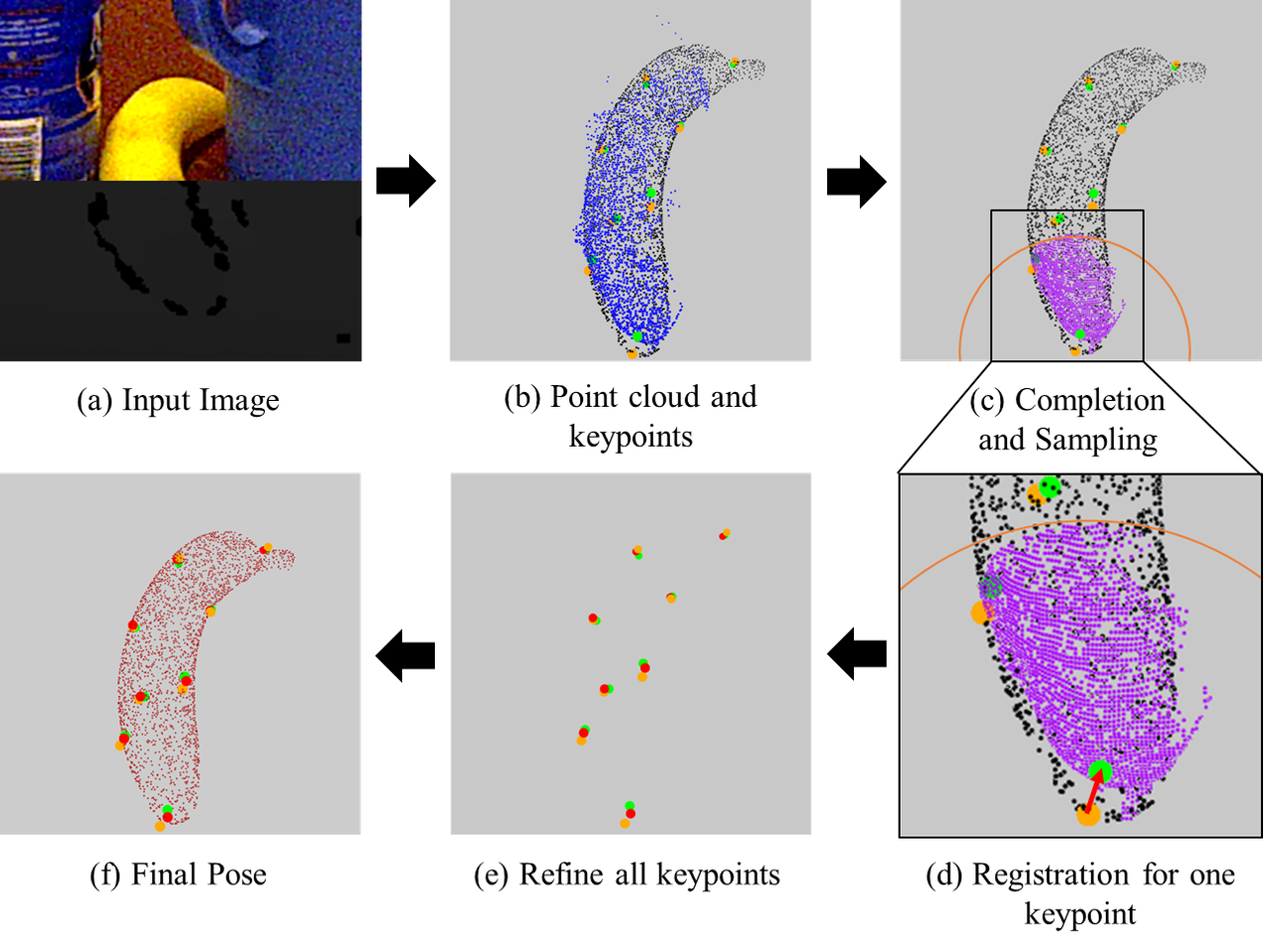}}
  \caption{\textbf{Steps of our method:} With input RGBD image (a) (the bottom half shows the depth map) and initial pose, we transform the visible point cloud (shown in blue, known object point cloud shown in black) and keypoints (shown in orange, groundtruth keypoints shown in green) to the object coordinate system (b). After completing the visible point cloud and sampling (purple points) around each keypoint within the sphere of radius $r$ (c), we iteratively register purple and black point cloud (d) and get all refined keypoints (shown in red) (e). Then, we use the least squares fitting method to get the final pose. The model transformed by the final pose is shown in (f). It is evident that the refined keypoints are closer to groundtruth than the original keypoints.}
  \label{fig:Pipeline}
  \end{figure}
  
Traditional methods\cite{brachmann2014learning, hinterstoisser2012model} attempted to extract hand-crafted features from the correspondences between known RGB images and object mesh models. However, these methods are less effective in heavy occlusion scenes or on low-texture objects. With the rapid development of deep learning, Deep Neural Networks (DNN) are now applied to the 6D object pose estimation task and demonstrate significant performance improvements. Specifically, some methods\cite{wang2019densefusion, zeng2019iros, xiang2017posecnn} use DNNs to directly regress the translation and rotation of each object. However, the non-linearity of the rotation results in poor generalization of these methods. More recently, works like\cite{peng2019pvnet, he2020pvn3d, zeng21icra} utilize DNNs to detect the keypoints of each object and subsequently compute the 6D pose parameters using Perspective-n-Point (PnP) for 2D keypoints or Least Squares methods for 3D keypoints. 
\IEEEpubidadjcol

While DNN methods can solve the problem more rapidly, they are still unable to achieve high accuracy due to errors in segmentation or regression. To achieve higher accuracy and stability, many works have adopted pose refinement methods, of which the most common is the Iterative Closest Point (ICP) \cite{besl1992method} algorithm. Given an estimated pose, the method tries to find the nearest neighbor of each point of the source point cloud in the target point cloud, considers it as the corresponding point, and solves for the optimal transformation iteratively. Moreover, works like \cite{li2018deepim, wang2019densefusion} use DNNs to extract more features for better performance. However, with the development of pose estimation networks, performance improvement of these pose refinement methods becomes less and less. The limited accuracy of existing registration methods can be attributed to their reliance on incomplete point clouds to register entire object mesh point clouds, resulting in numerous erroneous correspondences. \color{black}Besides, despite the widespread use of color information in 6D estimation, its potential to enhance registration accuracy remains largely unexplored. Conventional methods have not effectively exploited the benefits of color information and are primarily designed to solve the large-scale optimization problem of point cloud registration, rather than to deal with the small-scale problem of pose refinement, resulting in an untapped area of research.\color{black}
 
Our refinement method mainly contains two modules. Firstly, we propose a point cloud completion network to fully utilize the point cloud and RGB data. Our composite encoder of the network has two branches: the local branch fuses the RGB and point cloud information at each corresponding pixel, and the global branch extracts the feature of the whole point cloud. The decoder of the network follows \cite{yuan2018pcn} and employs a multistage point generation structure. Additionally, we add a keypoint detection module to the point cloud completion network during the training process to improve the sensitivity of the completed point cloud to pose accuracy, leading to better pose optimization. Secondly, to use color and point cloud data in registration and to enhance method stability, we propose a novel method named Color supported Iterative KeyPoint (CIKP), which samples the point cloud surrounding each key point and leverages both RGB and point cloud information to refine object keypoints iteratively. However, the CIKP method will make it hard to refine all key points when the point cloud is incomplete, which limits its performance. To address this issue, we introduce a combination of our completion network and the CIKP method, referred to as {\it Point Cloud Completion and Keypoint Refinement with Fusion} (PCKRF). \color{black}This integrated approach enables the refinement of the initial pose prediction from the pose estimation network. We further conduct extensive experiments on YCB-Video\cite{xiang2017posecnn} and Occlusion LineMOD\cite{brachmann2014learning} datasets to evaluate our method. The results demonstrate that our method can be effectively integrated with most existing pose estimation techniques, leading to improved performance in most cases.\color{black}

	\begin{figure*}[t]
		\centering
		\includegraphics[width=0.84\linewidth]{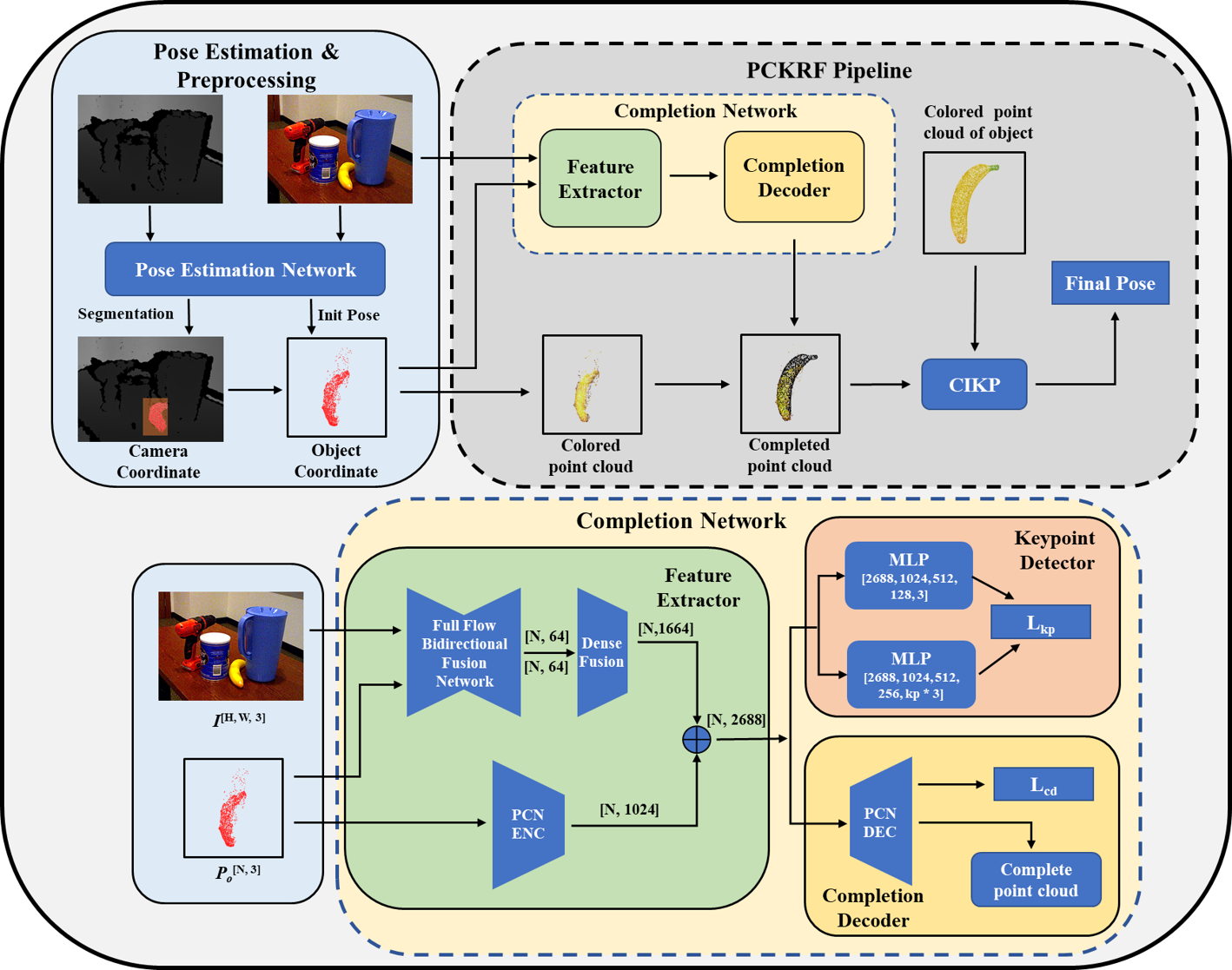}
		\caption{The upper diagram features the PCKRF pipeline and the lower diagram is the architecture of our point cloud completion network. In the preprocessing step, we utilize the segmentation result and pose of the target object given by the pose estimation network to obtain the partial point cloud in the object coordinate system. The PCKRF pipeline first completes the partial point cloud by the point completion network and then refines the initial pose by our CIKP method. In the point cloud completion network, the Feature Extractor fuses the point cloud and RGB color at each corresponding pixel, and the Keypoint Detector predicts the offset from each point to each keypoint to improve the sensitivity of the completed point cloud to pose accuracy. \color{black}The loss function of the completion network is a joint optimization of the keypoint detector Loss $L_{kp}$ and the completion decoder Loss $L_{cd}$. \color{black}}
		\label{fig:Inference}
	\end{figure*}	

Our main contribution is threefold:
\begin{itemize}
    \item PCKRF: A pipeline that combines our completion network and CIKP method, utilizing RGBD information and keypoints throughout the refinement. 
    \item A novel point completion network that includes a composite encoder and adds a keypoint detection module.
    \item A novel iterative pose refinement method CIKP that uses both RGB and point cloud information based on keypoints refinement.
\end{itemize}
\color{black}
Experiments demonstrate that our PCKRF exhibits superior stability compared to existing approaches when optimizing initial poses with relatively high precision. Notably, the results indicate that our method can be effectively integrated with most existing pose estimation techniques, leading to improved performance in most cases. Furthermore, our method achieves promising results even in challenging scenarios involving textureless and symmetrical objects.\color{black}

\section{Related Works}
\subsection{Pose Estimation}
Pose estimation methods can be categorized into two types based on their optimization goal: holistic and keypoint-based methods. Holistic methods predict the 3D position and orientation of objects directly from the provided RGB and/or depth images. Traditional template-based methods construct a rigid template for an object from different viewpoints and compute the best-matched pose for the given image \cite{cao2016real,hinterstoisser2011gradient}. Recently, some works utilized DNNs to directly regress or classify the 6D pose of objects. PoseCNN \cite{xiang2017posecnn} used a multi-stage network to predict pose. It first utilized Hough Voting to determine the center location of objects and then directly regressed 3D rotation parameters. SSD-6D \cite{kehl2017ssd} first detected objects in the images and then classified their poses. DenseFusion \cite{wang2019densefusion} fused RGB and depth values at the per-pixel level, which significantly impacted 6D pose estimation methods based on RGBD images. However, the non-linearity of the rotation makes it challenging for the loss function to converge. Recently, Neural Radiance Fields have also been employed for 6D pose estimation, showcasing significant inspiration and research potential\cite{lin2023parallel}.

\color{black}
Pose estimation using only point cloud information is also called point cloud registration. Recently, the advancements in deep neural networks, particularly in three-dimensional geometry with methods like PointNet \cite{Charles_Su_Kaichun_Guibas_2017} and DGCNN \cite{Wang_Sun_Liu_Sarma_Bronstein_Solomon_2019}, have significantly propelled the progress of deep point cloud registration. These methods are centered around the idea of utilizing deep neural networks to extract features from cross-source point clouds. These extracted features then serve as the basis for registrations or are directly used to regress transformation matrices. Techniques like SpinNet \cite{Ao_Hu_Yang_Markham_Guo_2021} aim to extract robust point descriptors through specialized neural network designs, focusing on feature learning. However, its reliance on a voxelization preprocessing step poses a challenge when dealing with cross-modality point clouds. Another approach, D3Feat \cite{Bai_Luo_Zhou_Fu_Quan_Tai_2020}, constructs features based on k-nearest neighbors. Nonetheless, this descriptor tends to struggle when confronted with significant density disparities. Beyond these point descriptor-centered methodologies, several strategies emphasize feature matching. For instance, Deep Global Registration (DGR) \cite{Choy_Dong_Koltun_2020} employs a UNet architecture to discern whether a point pair corresponds, reinterpreting the feature-matching challenge as a binary classification task. Alternatively, transformation learning approaches directly estimate transformations through neural networks. Feature-metric registration (FMR) \cite{Huang_Mei_Zhang_2020} introduces a technique that aligns two point clouds by minimizing their feature metric projection error, offering a unique approach to point cloud registration. More recently, attempts have been made to leverage the Transformer for aggregating context between two point clouds, followed by estimating correspondences through the utilization of dual normalization \cite{qin2022geometric} or some end-to-end pipelines without key points \cite{lee2021deeppro}. Another incremental method \cite{Zhang_2023_CVPR} that combined with deep-learned methods has also achieved excellent results. Moreover, the diffusion model is also applied to point cloud registration\cite{jiang2024se}. To further verify the effectiveness of our pipeline and its performance on texture-free objects, we selected a representative work \cite{qin2022geometric}, modified our framework, and conducted testing experiments using only the point cloud information.
\color{black}

Keypoint-based methods provide one way to address the above problems. Keypoint differs from superpoint \cite{yu2021cofinet,zhou2021patch2pix,qin2022geometric}, which relies on clustering or patching for point cloud registration without a prior model. Each keypoint is calculated based on the specific geometric features of the given model. YOLO-6D \cite{tekin2018real} employed the popular object detection model YOLO to predict 8 points and a center point for each bounding box of the object projected onto the 2D image. The method then computed the 6D pose using the PnP algorithm. PVNet \cite{peng2019pvnet} predicted a unit vector to each keypoint for each pixel, then voted the 2D location for each keypoint and calculated the final pose using the PnP algorithm. PVN3D \cite{he2020pvn3d} used additional depth information to detect 3D keypoints via Hough Voting and calculated the 6D pose parameters with the Least Squares method. In order to fully exploit the RGB and depth data, FFB6D \cite{he2021ffb6d} proposed a novel feature extraction network that applies fusion at both the encoding and decoding layers.

\subsection{Pose Refinement}
\color{black}
Most of the methods mentioned above apply pose refinement techniques to further improve the accuracy of their results. The most commonly used method is ICP \cite{besl1992method}, but it only leverages the Euclidean distance between points. Some methods \cite{chen1992object, segal2009generalized} tried to change the optimization goal to accelerate the iterative process or improve the result. Others \cite{johnson1999registration, men2011color, korn2014color} introduced a color space into ICP, which results in faster convergence and better performance than 3D ICP methods. There are also some works\cite{serafin2015nicp} that pay more attention to convergence speed and propose an Anderson acceleration approach\cite{zhang2021fast}.

The main difference between the registration and refinement methods is whether there is an initial registration pose. Some probabilistic point cloud registration methodologies can also be used for refinement. These approaches often leverage Gaussian Mixture Models (GMMs) to represent the distribution of point clouds, framing point cloud refinement as an optimization task involving probability density functions. Notably, GMM \cite{jian2010robust} emerges as a widely adopted method due to its robustness against considerable noise and outliers, despite its relatively higher computational demands. Alternatively, FilterReg \cite{Gao_Tedrake_2019} innovatively reformulates the correspondence challenge in point set registration as a filtering problem through the application of Gaussian filtering. However, these methods have higher adaptability to low-precision initial poses, but often perform poorly in further optimizing the accuracy, making them less suitable for high-precision 6D pose estimation problems.
\color{black}

Recently, DNN-based approaches were also used to address the refinement issue. For instance, given the initial pose and 3D object model, DeepIM\cite{li2018deepim} refines the pose iteratively by matching the rendered image with the observed image.  Manhardt et al. \cite{Manhardt_2018_ECCV} proposed a novel visual loss that refines the pose by aligning object contours with the initial pose. Densefusion\cite{wang2019densefusion} also proposed their refinement network, which follows their main network with the original RGB features and the corresponding features of the transformed point cloud as input. Considering 6D pose estimation methods and their impressive performance improvements, existing refinement methods may not always be able to maintain stability in the entire system and achieve the highest accuracy for existing pose estimation methods. To effectively optimize high-precision pose estimation, we propose a new pose estimation framework called PCKRF (Point Cloud Completion and Keypoint Refinement with Fusion Network), which relies on keypoints and point cloud completion.

\subsection{Point Cloud Completion}
\color{black}

VoxelNet \cite{zhou2018voxelnet} endeavors to segment the point cloud into voxel grids and utilizes convolutional neural networks, yet this voxelization process inevitably sacrifices intricate point cloud details. Additionally, enhancing the voxel grid's resolution leads to a substantial surge in memory usage. Yuan et al. \cite{yuan2018pcn} introduced PCN, a sophisticated approach rooted in PointNet \cite{Charles_Su_Kaichun_Guibas_2017} and FoldingNet \cite{yang2018foldingnet}, which employs a coarse-to-fine methodology. However, its decoder struggles to reconstitute uncommon object geometries, such as seat backs with gaps. Consequently, subsequent methods \cite{wang2020cascaded,xia2021asfm} have shifted their emphasis towards multi-step point cloud generation, facilitating the reconstruction of intricate details in the final point cloud. Furthermore, inspired by DGCNN \cite{Wang_Sun_Liu_Sarma_Bronstein_Solomon_2019}, several researchers have ventured into graph-based techniques \cite{wu2021point,zhu2021towards}, emphasizing regional geometric nuances. Recently, the transformer structure originated in the
field of natural language processing and has also been employed in addressing the issue of point cloud completion, yielding notably effective results \cite{zhou2022seedformer}.
\color{black}

\section{Our PCKRF Method}

Given an RGB image and/or a depth map, the task of 6D pose estimation is to predict the rigid transformation matrix $P \in SE(3)$ for each object in the image, which comprises a rotation matrix $R \in SO(3)$ and a translation matrix $T \in \mathbb{R}^3$. It transforms the object from its own coordinate system to the camera coordinate system. Given an observed RGB image and a depth map, we first obtain the predicted pose and segmentation result of an object from the pose estimation network. Then, we utilize our PCKRF pipeline to compute a relative transformation $\Delta P \in SE(3)$ for correcting the result. Specifically, we first complete the visible point cloud by our completion network, and then use the CIKP method to compute the refined pose. The summary of the inference process is shown in Fig. \ref{fig:Inference}.  

\subsection{Preprocessing}
\label{sub:preprocessing}
This part aims to obtain the keypoints, initial pose, and visible point cloud of each object. Firstly, we follow keypoint-based approaches \cite{he2021ffb6d, he2020pvn3d} to select $K$ keypoints on each object surface. Subsequently, the pose estimation network takes in an RGBD image as input and outputs the segmentation results and predicts the pose of each object in the image. Additionally, we convert the given segmented depth map $D^{H_s \times W_s }$ to the point cloud of the target object $P_c^{3 \times N}$, where $H_s, W_s$ are the height and weight of the segmented depth map, respectively, and $N$ is the number of points in the point cloud. Furthermore, we transform the point cloud from the camera coordinate system $P_c$ to the object coordinate system $P_o$ using the provided initial pose $R_{init} \in SO(3)$ and $T_{init} \in \mathbb{R}^3$ as follows:

\begin{equation}
    P_o = R_{init}^{-1} (P_c - T_{init}),
\end{equation}
Finally, we apply the MeanShift clustering algorithm \cite{cheng1995mean} to eliminate outliers, which enhances the subsequent performance of the point cloud registration.

\subsection{Completion Network for Pose Refinement}
\label{sub:network}
We improved the limited accuracy of existing refinement methods by incorporating a completion network as our pipeline's first module. This compensates for incomplete point clouds used to register entire object point clouds, which previously caused numerous erroneous correspondences. The architecture of our point cloud completion network is illustrated in Fig. \ref{fig:Inference}, which is composed of a feature extractor, a point cloud completion decoder, and a training-only keypoint detector.

\textbf{Feature Extractor:} To extract effective features from the partial point cloud and leverage the RGB and depth fusion data, we combine the Full Flow Bidirectional Fusion Network (FFB) \cite{he2021ffb6d}, Dense Fusion module (DF)\cite{wang2019densefusion} and PCN encoder \cite{yuan2018pcn} as Feature Extractor. The Feature Extractor includes two branches: the local feature extraction branch and the global feature extraction branch. After preprocessing, we get the object point cloud with the predicted pose in its coordinate system $P_o$. Then, $P_o$ and the observed RGB image $I^{H \times W \times 3}$ are fed into the FFB Fusion Network and DF module to get the local features of each point. 
Concretely, the FFB modules are applied to each representation learning layer as bridges for information communication between the RGB and point cloud feature extraction networks. Using the correspondence between individual points in the point cloud and pixels in the RGB image, the DF module fuses the two data sources and extracts pixel-wise dense features.
Simultaneously, the PCN encoder takes the partial point cloud as input and extracts the global features by stacking two PointNet\cite{qi2017pointnet} feature extraction modules.
Finally, local features and global features are concatenated as $F$.

\textbf{Decoder and Keypoint Detector:} In this section, we aim to generate a dense point cloud for the CIKP by implementing a keypoint detector with the completion network to enhance the sensitivity of the completed point cloud to pose accuracy. This is essential because CIKP requires high pose accuracy for the partial point clouds around each keypoint. Therefore, we incorporate a keypoint decoder and a PCN Decoder \cite{yuan2018pcn} to process the learned features $F$. The keypoint decoder predicts the keypoints and center offset, while the PCN decoder completes the visible point cloud. Given a set of keypoints $P_k^{3 \times K}$ and a visible point cloud $P_o$, we define the translation offset from the $i_{th}$ point $p_i \in P_o$ to the $j_{th}$ keypoint $kp_j \in P_k$ as $of_i^j$. We supervise the keypoint detector module using $L_{kp}$ loss:

\begin{equation}
    L_{kp} = \frac{1}{N} \sum_{i=1}^{N} \sum_{j=1}^{K} ||of_i^j - of_i^{j*} ||,
\end{equation}
where $K$ is the number of keypoints, $of_i^{j*}$ is the groundtruth of translation offset. Note that $K = 1$ if $L_{kp}$ is used for predicting center point offset. We denote the loss of keypoints offset and center offset as $L_{kp}$ and $L_c$, respectively.

\begin{equation}
    L_{c} = \frac{1}{N} \sum_{i=1}^{N} ||of_i^j - of_i^{j*} ||
\end{equation}

We supervise the completion decoder using $L_{cd}$ loss:

\begin{equation}
    L_{cd} = \frac{1}{|P_o|} \sum_{x \in P_o} \min_{y \in P_d} ||x - y||_2 + 
    \frac{1}{|P_d|} \sum_{y \in P_d} \min_{x \in P_o} ||x - y||_2,
    \label{equ:loss_cd}
\end{equation}
where $P_d$ is the output dense point cloud. The first term of the formula represents the sum of the minimum distance from any point $x$ in $P_o$ to $P_d$, while the second term represents the minimum distance from any point $y$ in $P_d$ to $P_o$. The overall loss $L$ is calculated as below, where $\alpha, \beta, \gamma$ represent the weights of different losses:

\begin{equation}
\label{loss}
    L = \alpha L_{kp} + \beta L_c + \gamma L_{cd},
\end{equation}
\color{black}
By treating the positions of keypoints as an optimization objective, we aim to encourage the neural network to extract more features in their vicinity and to focus on the information surrounding the keypoints. As keypoints represent critical locations of an object, optimizing their positions can be viewed as a way to enhance their completion weights.
\color{black}
\subsection{Color Supported Iterative KeyPoint}
Traditional registration algorithms for pose estimation only utilize the Euclidean distance between the source point cloud and the target point cloud. These methods are less stable since they only consider the registration of the entire point cloud without color information. In contrast, pose estimation networks that fuse color and point cloud information demonstrate notable improvement in estimation accuracy. Motivated by this, we propose CIKP that iteratively refines each keypoint using the position and color information of each point.

We first define the distance between two colored points as in Eq. (\ref{equ:drgb}). Given two colored points $p_1$ and $p_2$, we divide them into a position component $x_1, x_2 \in \mathbb{R}^3$ and a color component $c_1, c_2 \in \mathbb{R}^3$ respectively. Note that $c_1, c_2 \in [0, 1]^3$. The final distance between $p_1, p_2$ is their distance in the Euclidean space plus the distance in weighted color space. We stipulate that the closer the spatial distance, the higher the weight of the color distance. To mitigate the issue of excessively large weights resulting from small spatial distances, we implement a minimum weight value of 1. In addition, we introduce a threshold parameter, denoted as $\epsilon$, which determines the threshold distance between two points. If the distance between two points is below $\epsilon$, the weight assigned to them will remain at 1. By adjusting the value of $\epsilon$, we can effectively control the impact of the color information on the overall computation process.

\begin{equation}
\label{equ:drgb}
D = D_1 + w D_2,\  \mbox{where}\left \{
\begin{aligned}
&D_1 = ||x_1 - x_2||_2, \\
&D_2 = ||c_1 - c_2||_2, \\
&w = \min (\frac{\epsilon}{D_1}, 1)
\end{aligned}
\right .
\end{equation}

We summarize the process of CIKP in Alg. \ref{alg:cikp}. The source point cloud $P_s$ is a colored point cloud of objects, and the target point cloud $P_t$ consists of visible colored point clouds and uncolored point clouds completed by our completion network, where both $P_s$ and $P_t$ are in the object coordinate system. We first select a keypoint, take it as the center, and collect all points in the $P_t$ within the sphere of radius $r$, denoted by $S_p$. If $|S_p|$ is less than a threshold $m_1$, we keep its original state and select the next keypoint. If $|S_p|$ is more than $m_2$, we randomly select $m_2$ points in $S_p$. Then, for each $p_t \in S_p$, we find its closest point in $P_s$. If $p_t$ is colored, we use Eq. \eqref{equ:drgb} to calculate distance, otherwise we calculate their Euclidean distance directly. After that, we calculate the optimal translation transformation $T_k$ between the two point clouds and transform the selected keypoint with $T_k$. After all keypoints have been transformed, the refined pose is calculated using the Least Squares method. We repeat the above steps until the mean distance between corresponding points is less than a threshold $\tau$ or reaches the maximum number of iterations. Note that we decouple translation and rotation here. Only translation is taken into account when calculating the transformation of each keypoint. Rotation is only considered after all the keypoints are optimized. This approach helps to mitigate the risk of local overfitting, which will be demonstrated in subsequent experiments.

\begin{algorithm}
\caption{Color Supported Iterative KeyPoint}
\label{alg:cikp}
\LinesNumbered
\SetKwInOut{Init}{Initialize}
\KwIn{source point cloud $P_s$, target point cloud $P_t$, keypoint set $S_k$, search radius $r$, threshold $m_1, m_2$}
\KwOut{refined rotation $R$ and translation $T$}
\Init {sampled point cloud set $S_{p}$, closest point set $M$, refined keypoint set $S_{kpr}$, source point cloud in camera coordinate system $P_a$}
$T \leftarrow [0 \ 0 \ 0]^T$

$R \leftarrow I^{3 \times 3}$

\While{ not converged}{
    $P_a \leftarrow R \times P_s + T$
    
    $S_{kpr} \leftarrow R \times S_k + T$
    
    \ForEach{ $i \leftarrow 1 : n_{kp} $}{
        $S_p \leftarrow \varnothing$
        
        $M \leftarrow \varnothing$
        
        \ForEach{$p_t \in P_t$}{
            \If{$distXYZ(S_{kpr}[i], p_t) < r$}{
                push $p_t$ into $S_p$
            }
        }
        \If{$|S_p| < m_1$}{ \textbf{continue}}
        \If{$|S_p| > m_2$}{ Random select $m_2$ points in $S_p$}
        \ForEach{$p_t \in S_p$}{
        $p \leftarrow FindClosestPointInP_a(p_t)$
        
        push $p$ into $M$
        }
        
        $T_k \leftarrow  \underset{T_k}{\arg\min} \  \underset{j}{\sum}distXYZ(S_p[j],M[j] + T_k) $
        
        $S_{kpr}[i] \leftarrow S_{kpr}[i] + T_k$
    }
    $[R, T] = \underset{R, T}{\arg\min} \ \underset{i}{\sum} distXYZ(R \times S_k[i] + T, S_{kpr}[i]) $

}
\end{algorithm}

The poses $R$ and $T$ calculated in this part represent the relative transformations from the source point cloud to the target point cloud in the object coordinate system. However, we need to perform a final conversion step to obtain the final pose estimation result, which is transforming the object from the object coordinate system to the camera coordinate system. The initial rotation and translation are denoted as $R_{init}$ and $T_{init}$ respectively. Then, the initial pose $\theta_{init}$ and the calculated relative pose $\Delta\theta$ can be represented as follows:

\begin{align}
    \theta_{init} &= 
    \left[
    \begin{array}{cc}
        R_{init}^{3 \times 3}& T_{init}^{3 \times 1} \\
        \mathbf{0}^{1 \times 3}& 1 
    \end{array}
    \right] \\
    \Delta\theta &=
    \left[
    \begin{array}{cc}
        R^{3 \times 3}& T^{3 \times 1} \\
        \mathbf{0}^{1 \times 3}& 1
    \end{array}
    \right] 
\end{align}

the final refined pose can be calculated as follows:
\begin{equation}
\begin{aligned}
    \theta &= \theta_{init} \times \Delta\theta \\
    &=     
    \left[
    \begin{array}{cc}
        R_{init} \times R & R_{init} \times T + T_{init} \\
        \mathbf{0}^{1 \times 3}& 1
    \end{array}
    \right] 
\end{aligned}
\end{equation}

\section{Experiment}
\subsection{Datasets}

\begin{table*}[h]
    \centering
    
    \color{black}
    \caption{Result of 6D Pose Estimation on the YCB-Video Dataset using FFB6D output as initial pose. The ADD-S and ADD(S) AUC are reported. Objects with bold names are symmetric. Bold values are the highest score. Underline values indicate results lower than the initial value.}
    \resizebox{\textwidth}{!}{
    \begin{tabular}{l|c c|c c|c c|c c|c c|c c}
    \hline
        &
        \multicolumn{2}{c|}{FFB6D\cite{he2021ffb6d}}&
        \multicolumn{2}{c|}{+GeoTransformer\cite{qin2022geometric}}&
        \multicolumn{2}{c|}{+ICP-plane\cite{chen1992object}}& 
        \multicolumn{2}{c|}{+GICP\cite{segal2009generalized}}&
        \multicolumn{2}{c|}{+Colored 6D \cite{johnson1999registration}}&
        \multicolumn{2}{c}{+Ours}\\
        \hline
        &
        ADD-S & ADD(S) & ADD-S & ADD(S) & ADD-S & ADD(S) & ADD-S & ADD(S) & ADD-S & ADD(S) & ADD-S & ADD(S)\\
        \hline 
        master\_chef\_can & 96.4  & 80.7  & \underline{95.4}  & \underline{80.1}  & \underline{95.1}  & \underline{76.9}  & \underline{95.3}  & \underline{79.1}  & \underline{95.5}  & \underline{80.1}  & \textbf{96.5}  & \textbf{81.7}  \\ 
        cracker\_box & 96.4  & 95.0  & \underline{94.8}  & \underline{93.2}  & 96.4  & 95.4  & \underline{96.3}  & \underline{94.9}  & \underline{94.9}  & \underline{94.1}  & \textbf{96.7}  & \textbf{95.6}  \\ 
        sugar\_box & 97.7  & 96.8  & \underline{95.7}  & \underline{94.9}  & \underline{97.6}  & \underline{96.7}  & 97.8  & 97.1  & \underline{97.6}  & 97.0  & \textbf{98.0}  & \textbf{97.3}  \\ 
        tomato\_soup\_can & 95.8  & 88.1  & \underline{95.2}  & \underline{88.0}  & \textbf{95.9}  & \underline{82.4}  & \underline{95.6}  & \underline{84.3}  & \underline{95.5}  & \underline{87.4}  & \textbf{95.9}  & \textbf{88.3}  \\ 
        mustard\_bottle & 98.1  & 97.6  & 98.3  & 98.0  & 98.1  & 97.6  & 98.1  & 97.8  & 98.2  & \textbf{98.0}  & \textbf{98.4}  & \textbf{98.0}  \\ 
        tuna\_fish\_can & 97.2  & 91.3  & \underline{96.5}  & \underline{91.1}  & \underline{96.7}  & \underline{88.7}  & \underline{96.3}  & \underline{86.6}  & 97.2  & 91.6  & \textbf{97.3}  & \textbf{92.0}  \\ 
        pudding\_box & 96.3  & 93.1  & \underline{94.9}  & \underline{91.3}  & \underline{95.0}  & \underline{90.4}  & \underline{96.1}  & \underline{92.6}  & \underline{95.1}  & \underline{90.6}  & \textbf{96.7}  & \textbf{94.2}  \\ 
        gelatin\_box & 97.8  & 95.8  & 97.8  & 95.8  & 97.8  & \underline{95.0}  & \textbf{98.3}  & \textbf{97.0}  & 97.8  & \underline{95.0}  & 98.1  & 96.2  \\ 
        potted\_meat\_can & 92.6  & 89.8  & \underline{91.2}  & \underline{88.9}  & 92.6  & \underline{88.9}  & \underline{92.4}  & \underline{88.2}  & \underline{92.3}  & \underline{87.9}  & \textbf{92.9}  & \textbf{90.0}  \\ 
        banana & 97.4  & 94.9  & \underline{96.8}  & 94.1  & 97.9  & 96.0  & \textbf{98.2}  & \textbf{97.1}  & 97.8  & 96.0  & 97.8  & 95.8  \\ 
        pitcher\_base & 97.7  & 97.0  & \underline{95.5}  & \underline{93.6}  & 97.9  & 97.4  & \textbf{98.0}  & \textbf{97.6}  & 97.9  & 97.4  & 97.8  & 97.2  \\ 
        bleach\_cleanser & 96.5  & 93.7  & 96.9  & 94.5  & \textbf{97.3}  & \textbf{96.2}  & \textbf{97.3}  & 96.0  & 96.9  & 95.2  & 96.9  & 94.5  \\ 
        \textbf{bowl} & 95.8  & 95.8  & 97.1  & 97.1  & 96.6  & 96.6  & \textbf{97.3}  & \textbf{97.3}  & 96.5  & 96.5  & 96.6  & 96.6  \\ 
        mug & 97.5  & 95.3  & \underline{96.8}  & \underline{95.7}  & 97.7  & \underline{95.1}  & \textbf{97.8}  & 95.7  & 97.6  & 95.4  & \textbf{97.8}  & \textbf{96.2}  \\ 
        power\_drill & 97.3  & 96.2  & 97.9  & 96.7  & 97.8  & 97.2  & \textbf{98.0}  & \textbf{97.4}  & 97.6  & 96.8  & 97.8  & 97.2  \\ 
        \textbf{wood\_block} & 93.1  & 93.1  & \underline{92.7}  & \underline{92.7}  & 94.4  & 94.4  & \textbf{95.2}  & \textbf{95.2}  & 94.5  & 94.5  & 94.3  & 94.3  \\ 
        scissors & \textbf{98.1}  & \textbf{97.1}  & \underline{96.1}  & \underline{93.6}  & \underline{97.2}  & \underline{95.2}  & \underline{96.3}  & \underline{93.5}  & \underline{95.9}  & \underline{95.6}  & \textbf{98.1}  & \underline{97.0}  \\ 
        large\_marker & 96.9  & 90.0  & \underline{96.6}  & \underline{88.5}  & 98.0  & \underline{89.9}  & \textbf{98.1}  & \underline{88.4}  & 97.8  & \underline{89.8}  & 97.8  & \textbf{90.5}  \\ 
        \textbf{large\_clamp} & \textbf{96.8}  & \textbf{96.8}  & \underline{96.7}  & \underline{96.7}  & \underline{95.8}  & \underline{95.8}  & \underline{95.4}  & \underline{95.4}  & \underline{95.8}  & \underline{95.8}  & \underline{96.7}  & \underline{96.7}  \\ 
        \textbf{extra\_large\_clamp} & \textbf{96.1}  & \textbf{96.1}  & \underline{96.0}  & \underline{96.0}  & \underline{94.1}  & \underline{94.1}  & \underline{93.3}  & \underline{93.3}  & \underline{95.4}  & \underline{95.4}  & \underline{95.6}  & \underline{95.6}  \\ 
        \textbf{foam\_brick} & 97.6  & 97.6  & \underline{97.4}  & \underline{97.4}  & \textbf{97.8}  & \textbf{97.8}  & 97.6  & 97.6  & \underline{97.3}  & \underline{97.3}  & 97.7  & 97.7  \\ \hline
        ALL & 96.6  & 92.9  & \underline{96.0}  & 93.2  & \underline{96.5}  & \underline{91.9}  & \underline{96.4}  & \underline{92.1}  & \underline{96.3}  & \underline{92.8}  & \textbf{96.8}  & \textbf{93.4} \\   \hline
    \end{tabular}
}
    \label{tab:ycb_comparison1}
\end{table*}

\color{black}

We evaluate our proposed method on two benchmark datasets.

\paragraph{YCB-Video} The YCB-Video dataset consists of 21 objects selected from the YCB object set. It contains 92 videos captured by RGBD cameras, and each video consists of 3-9 objects, leading to a total of over 130K frames. We followed previous works \cite{xiang2017posecnn,he2021ffb6d,wang2019densefusion} to split them into the training and testing sets. The training set also includes 80,000 synthetic images released by \cite{xiang2017posecnn}.

\paragraph{Occlusion LineMOD} The Occlusion LineMOD dataset \cite{brachmann2014learning} is re-annotated from the Linemod \cite{hinterstoisser2012model} dataset to compensate for its lack of occlusion. Unlike the LineMOD dataset, each frame in the Occlusion LineMOD dataset contains multiple heavily occluded objects, making it more challenging. We follow the previous work \cite{xiang2017posecnn} to split the training and testing sets and generate synthetic images for training.

\subsection{Evaluation Metrics}
We use the average distance (ADD) and average distance for symmetric objects (ADD-S) as metrics. Given the set of object point cloud $\mathcal{O}$ with $m$ points, the ground truth pose $[R^*, T^*]$ and predicted pose $[R, T]$, ADD, and ADD-S are defined as follows:
\begin{equation}
    ADD = \frac{1}{m} \underset{x \in \mathcal{O}}{\sum} ||(Rx + T) - (R^*x + T^*)||,
\end{equation}

\begin{equation}
    ADD\text{-}S = \frac{1}{m} \underset{x_1 \in \mathcal{O}}{\sum} \underset{x_2 \in \mathcal{O}}{\min}  ||(Rx_1 + T) - (R^*x_2 + T^*)||,
\end{equation}

In the YCB-Video dataset, we report the area under ADD-S and ADD(S) curve (AUC) following \cite{xiang2017posecnn} and set the maximum threshold of success cases to be 0.1m. The ADD(S) computes ADD-S for symmetric objects and ADD for others. In the Occlusion LineMOD dataset, we report the accuracy of ADD(S) with distance less than 10\% (ADD(S)-0.1) and 5\% (ADD(S)-0.05) of the diameter of the objects. 

\subsection{Implementation Details}
\label{sub:details}
All experiments were conducted on a PC with an Intel E5-2640-v4 CPU and NVIDIA RTX2080Ti GPU. For the segmentation results and initial poses required as inputs, we utilize pre-trained FFB6D\cite{he2021ffb6d} and PVN3D\cite{he2020pvn3d} results. In Full Flow Bidirectional Fusion Network, a PSPNet \cite{zhao2017pyramid} with a ResNet34 \cite{he2016deep} pre-trained on ImageNet \cite{deng2009imagenet} is applied to extract RGB image features. For each object, We randomly sample 2,048 points and apply RandLA-Net \cite{hu2020randla} to extract the point cloud geometry features. These features are then fused by DenseFusion \cite{wang2019densefusion}. The PCN-ENC block consists of two stacks of PointNet to extract the point cloud geometry features. The keypoint detector block consists of two MLPs whose details are shown in Fig. \ref{fig:Inference}. We follow \cite{yuan2018pcn} to employ a multi-stage structure in PCN-DEC to output a coarse point cloud (2,048 points) and a detailed point cloud (8,192 points). \color{black}We set $\alpha = \beta = 1, \gamma = 10$ in Eq. (\ref{loss}) and search radius $r$ to be 0.8 times the radius of the selected object. For the keypoints, we apply the SIFT-FPS \cite{he2021ffb6d} algorithm to select $K=8$ keypoints for each target object. We set threshold $m_1 = 100$ and $m_2 = 3000$ in Alg. \ref{alg:cikp}, we also set the tolerance to 0.00001 and max iteration number to 50. For the threshold $\epsilon$ in Eq. \eqref{equ:drgb}, we set $\epsilon = 5e^{-3}$, This means that if the Euclidean distance between two points is less than or equal to 5mm, the color distance weight will always be equal to the space distance weight.\color{black}

\subsection{Evaluation on the YCB-Video Dataset}

\begin{table*}[h]
    \centering
    \color{black}
    \caption{Result of 6D Pose Estimation on the YCB-Video Dataset using PVN3D output as initial pose.}
    \resizebox{\textwidth}{!}{
    \begin{tabular}{l|c c|c c|c c|c c|c c|c c}
    \hline
        &
        \multicolumn{2}{c|}{PVN3D\cite{he2020pvn3d}}&
        \multicolumn{2}{c|}{+GeoTransformer\cite{qin2022geometric}}&
        \multicolumn{2}{c|}{+ICP-plane\cite{chen1992object}}& 
        \multicolumn{2}{c|}{+GICP\cite{segal2009generalized}}&
        \multicolumn{2}{c|}{+Colored 6D \cite{johnson1999registration}}&
        \multicolumn{2}{c}{+Ours}\\
        \hline
        &
        ADD-S & ADD(S) & ADD-S & ADD(S) & ADD-S & ADD(S) & ADD-S & ADD(S) & ADD-S & ADD(S) & ADD-S & ADD(S)\\
        \hline 
        master\_chef\_can & 96.0  & 80.8  & \underline{95.1}  & \underline{80.1}  & \underline{95.0}  & \underline{77.6}  & \underline{95.1}  & \underline{79.4}  & \textbf{96.1}  & \textbf{81.1}  & \underline{95.7}  & \underline{80.4}  \\ 
        cracker\_box & 96.0  & 94.5  & \underline{93.8}  & \underline{93.4}  & \textbf{96.4}  & \textbf{95.3}  & 96.3  & 94.9  & 96.2  & 94.9  & 96.3  & 95.2  \\ 
        sugar\_box & 97.1  & 95.5  & 97.1  & 95.8  & 97.6  & 96.7  & \textbf{97.6}  & \textbf{96.9}  & 97.4  & \underline{96.1}  & \textbf{97.6}  & \textbf{96.9}  \\ 
        tomato\_soup\_can & 95.5  & 88.4  & \underline{95.1}  & \underline{88.1}  & \textbf{95.8}  & \underline{82.8}  & 95.5  & \underline{83.8}  & 95.5  & 88.5  & \textbf{95.8}  & \textbf{88.9}  \\ 
        mustard\_bottle & 97.8  & 97.0  & \underline{96.9}  & \underline{96.7}  & \textbf{98.1}  & 97.6  & \textbf{98.1}  & \textbf{97.8}  & 97.9  & 97.2  & 97.9  & 97.2  \\ 
        tuna\_fish\_can & 96.3  & 90.1  & \underline{94.9}  & \underline{90.0}  & 96.8  & \underline{88.5}  & 96.4  & \underline{87.2}  & 96.3  & 90.2  & \textbf{97.0}  & \textbf{90.4}  \\ 
        pudding\_box & 96.9  & 95.3  & \underline{94.9}  & \underline{91.3}  & \underline{94.9}  & \underline{90.4}  & \underline{95.9}  & \underline{92.1}  & \textbf{97.2}  & \textbf{95.8}  & \textbf{97.2}  & 95.4  \\ 
        gelatin\_box & 97.8  & 96.2  & \underline{97.7}  & \underline{95.8}  & \underline{97.7}  & \underline{94.9}  & \textbf{98.2}  & \textbf{97.0}  & 97.8  & 96.3  & \textbf{98.2}  & 96.3  \\ 
        potted\_meat\_can & 93.0  & 88.6  & \underline{91.2}  & 88.9  & \underline{92.8}  & \underline{88.3}  & \underline{92.6}  & \underline{87.7}  & \underline{92.9}  & \underline{88.4}  & \textbf{93.1}  & \textbf{88.9}  \\ 
        banana & 96.5  & 93.5  & 96.8  & 94.5  & 97.9  & 96.0  & \textbf{98.1}  & \textbf{96.9}  & 96.9  & 94.3  & \textbf{98.1}  & 96.5  \\ 
        pitcher\_base & 96.9  & 95.6  & \underline{95.8}  & \underline{94.8}  & 97.9  & 97.4  & \textbf{98.0}  & \textbf{97.5}  & 97.1  & 96.1  & 97.5  & 96.6  \\ 
        bleach\_cleanser & 96.3  & 93.6  & 96.4  & 94.3  & \textbf{97.3}  & \textbf{96.2}  & 97.2  & 95.9  & 96.6  & 94.4  & 96.8  & 94.7  \\ 
        \textbf{bowl} & 89.4  & 89.4  & \textbf{94.1}  & \textbf{94.1}  & 92.4  & 92.4  & 90.9  & 90.9  & 91.5  & 91.5  & 92.8  & 92.8  \\ 
        mug & 97.4  & 95.0  & \underline{96.7}  & 95.5  & 97.8  & 95.5  & 97.8  & 95.7  & 97.5  & 95.4  & \textbf{97.9}  & \textbf{96.0}  \\ 
        power\_drill & 96.6  & 95.1  & 96.8  & 95.7  & 97.8  & 97.2  & \textbf{98.0}  & \textbf{97.4}  & 97.0  & 95.8  & 97.5  & 96.6  \\ 
        \textbf{wood\_block} & 90.4  & 90.4  & 91.8  & 91.8  & \textbf{93.9}  & \textbf{93.9}  & 93.4  & 93.4  & 90.5  & 90.5  & 91.5  & 91.5  \\ 
        scissors & 96.5  & 92.7  & \underline{93.1}  & \underline{90.3}  & \textbf{97.2}  & \textbf{95.1}  & \underline{95.4}  & \underline{91.4}  & 96.8  & 93.2  & \underline{96.0}  & 92.7  \\ 
        large\_marker & 96.8  & 91.7  & \underline{96.5}  & \underline{88.4}  & 98.2  & 91.4  & 98.3  & 91.3  & 97.2  & 92.1  & \textbf{98.4}  & \textbf{92.7}  \\ 
        \textbf{large\_clamp} & 90.5  & 90.5  & 91.7  & 91.7  & \textbf{93.0}  & \textbf{93.0}  & 92.6  & 92.6  & 91.7  & 91.7  & 92.3  & 92.3  \\ 
        \textbf{extra\_large\_clamp} & 87.1  & 87.1  & 90.0  & 90.0  & \textbf{90.9}  & \textbf{90.9}  & 89.6  & 89.6  & 88.8  & 88.8  & 89.6  & 89.6  \\ 
        \textbf{foam\_brick} & 96.8  & 96.8  & \underline{96.4}  & \underline{96.4}  & \textbf{97.6}  & \textbf{97.6}  & 97.1  & 97.1  & 96.7  & 96.7  & 97.3  & 97.3  \\ \hline
        ALL & 95.2  & 91.5  & \underline{94.9}  & 92.3  & \textbf{95.9}  & 91.6  & \textbf{95.9}  & 91.6  & 95.6  & 92.0  & \textbf{95.9}  & \textbf{92.5} \\    \hline
    \end{tabular}
}
    \label{tab:ycb_comparison2}
\end{table*}

\color{black}

Table \ref{tab:ycb_comparison1} shows the results for all the 21 objects on the YCB-Video dataset using the pre-trained FFB6D\cite{he2021ffb6d} model output as the initial pose. Additionally, Table \ref{tab:ycb_comparison2} shows the results 
when the pretrained PVN3D\cite{he2020pvn3d} model output is the initial pose. We report the ADD-S AUC and ADD(S) AUC metrics for four ICP variants and our proposed method. 

\color{black}
The results in Table \ref{tab:ycb_comparison1} show that when FFB6D output is the initial pose, the point-to-plane ICP method \cite{chen1992object} and the GICP method \cite{segal2009generalized} yield inferior results compared to the initial ones. Additionally, all ICP variants are inferior to the initial results on almost half of the objects, which confirms that the classic ICP methods make it difficult to optimize the existing high-precision pose estimation method. Besides, the learning-based approach GeoTransformer \cite{qin2022geometric} is also inferior to the initial pose in most cases.
In contrast, our method only shows slightly inferior results in scissors, large clamp, and extra-large clamp, where two of the objects are only 0.1\% lower than the initial results. It is worth noting that the point-to-plane ICP and GICP methods show better results in ADD-S than in ADD(S), which demonstrates that these two methods encounter a significant amount of mismatch in establishing correspondence between the two point clouds. The reason behind this is the use of point-to-plane or plane-to-plane registration methods, which converge quickly in large-scale point cloud registration. However, these methods are more prone to mismatches when solving high-precision registration of small objects.

The results in Table \ref{tab:ycb_comparison2} show that when PVN3D output is used as the initial pose, our method also outperforms the initial results in ADD-S by 0.3\% and is one of the highest results among all methods. Compared with FFB6D, all ICP methods improve the original results when PVN3D output is used as the initial pose. However, point-to-plane ICP and GICP are still inferior to the initial results on 6 objects in ADD(S), whereas Colored 6D ICP and ours are only inferior to the initial results on 2 objects each. With the ADD(S) evaluation metric, GeoTransformer exhibits inferior performance compared to the initial results on 13 objects. The learning-based method GeoTransformer demonstrates poor performance when utilizing the initial poses from PVN3D and FFB6D. We believe this is primarily due to the fact that learning-based methods operate as black boxes, which are primarily designed for registration rather than refinement. Additionally, the lack of adjustable parameters specific to the task contributes to their suboptimal performance. These results reinforce the stability of our proposed method.

\color{black}

\begin{figure}[h]
    \centering
    \includegraphics[width=\linewidth]{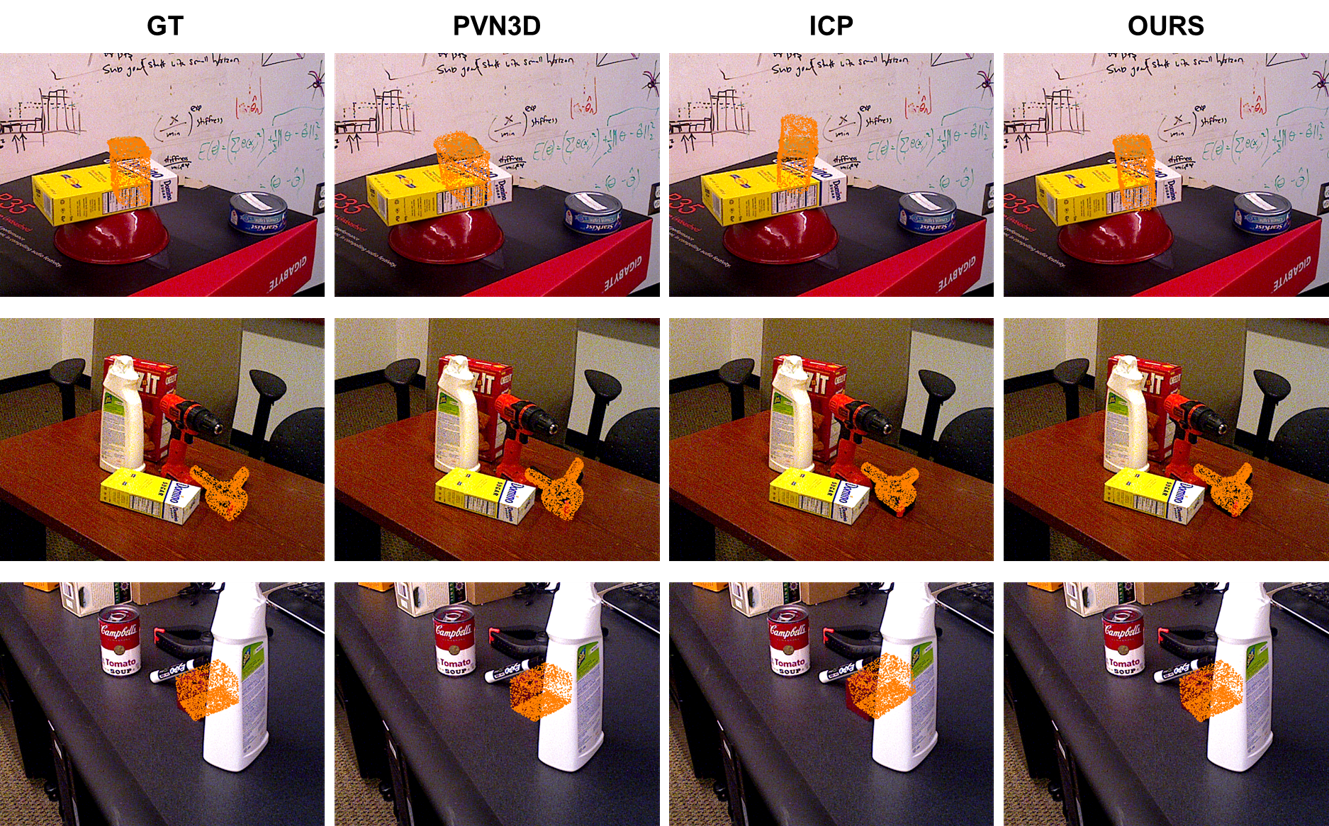}
    \caption{Qualitative results on the YCB-Video Dataset. Our method can outperform other methods with both low-precision initial pose (row 1) and high-precision initial pose (rows 2 and 3) with higher accuracy.}
    \label{fig:vis1}
\end{figure}

Figure \ref{fig:vis1} shows some of the qualitative test results using the PVN3D output as the initial pose. The figure reveals that our proposed method produces a more stable result, with a smaller change in the initial pose compared to the ICP method. Furthermore, the proposed method's performance improves with the assistance of the completed point cloud in scenes with severe occlusion, leading to excellent results.

\subsection{Evaluation on the Occlusion LineMOD Dataset}

\begin{table*}[h]
    \centering
    \color{black}
    \caption{Result of 6D Pose Estimation on Occlusion LineMOD Dataset with FFB6D output as initial pose. The ADD(S)-0.1 and ADD(S)-0.05 metrics are reported. Objects with bold names are symmetric. Bold values are the highest score. Underline values indicate results lower than the initial value.}
    \begin{tabularx}{0.9\linewidth}{l*{6}{|>{\centering\arraybackslash}X >{\centering\arraybackslash}X}}
    \hline
        &
        \multicolumn{2}{c|}{FFB6D\cite{he2021ffb6d}}&
        \multicolumn{2}{c|}{+GeoTransformer\cite{qin2022geometric}}&
        \multicolumn{2}{c|}{+ICP-plane\cite{chen1992object}}& 
        \multicolumn{2}{c|}{+GICP\cite{segal2009generalized}}&
        \multicolumn{2}{c|}{+Colored 6D \cite{johnson1999registration}}&
        \multicolumn{2}{c}{+Ours}\\
        \hline
        ADD(S)-N & 0.1 & 0.05 & 0.1 & 0.05 & 0.1 & 0.05 & 0.1 & 0.05 & 0.1 & 0.05 & 0.1 & 0.05\\
        \hline
        ape & 55.0 & 20.0 & 55.3 & 21.2 & 57.2 & \underline{18.4} & 57.2 & \underline{19.7} & 57.3 & \underline{19.7}  & \textbf{57.6}  & \textbf{21.9} \\ 
        can & 82.5 & 56.3 & \underline{82.1} & \textbf{65.7} & \underline{81.6} & 60.4 & \underline{81.6} & 60.6 & \underline{81.7} & 63.1  & \textbf{83.4}  & 64.5 \\ 
        cat & 37.1 & 14.1 & 37.9 & 18.3 & \underline{37.0} & 16.5 & \underline{36.9} & 16.1 & \underline{36.9} & 16.2  & \textbf{38.3}  & \textbf{22.3} \\ 
        driller & 76.5 & 50.7 & 80.3 & 67.3 & 83.4 & 71.6 & \textbf{84.3} & \textbf{73.6} & 81.8 & 68.3  & 79.8  & 61.4 \\
        duck & 56.6 & 13.6 & 56.9 & 17.3 & 60.7 & 14.9 & 60.3 & 16.6 & 60.4 & 17.0  & \textbf{61.4}  & \textbf{17.9} \\ 
        \textbf{eggbox} & 59.1 & 19.7 & 60.6 & 24.8 & \underline{58.2} & 24.6 & \underline{57.4} & \textbf{26.1} & \underline{57.6} & 25.1  & \textbf{59.5}  & 24.1 \\ 
        \textbf{glue} & 66.7 & 53.8 & \underline{65.8} & \underline{52.1} & \underline{64.1} & \underline{53.3} & \underline{64.0} & \underline{52.9} & \underline{63.9} & \underline{50.7}  & \textbf{66.9}  & \textbf{57.7} \\ 
        holepuncher & 84.6 & 41.4 & \underline{83.5} & \textbf{57.6} & 85.3 & 49.6 & 85.6 & 53.4 & 84.6 & 57.2  & \textbf{87.4}  & 54.6 \\ 
        \hline
        Average & 64.8  & 33.7  & 65.3  & 39.7  & 65.9 & 38.7  & 65.9 & 39.9  & 65.5  & 39.7  & \textbf{66.8}  & \textbf{40.6} \\ 
        \hline
    \end{tabularx}
    \label{tab:occ_comparison1}
\end{table*}
\color{black}

\begin{table*}[h]
    \color{black}
    \centering
    \caption{Result of 6D Pose Estimation on Occlusion LineMOD Dataset using PVN3D output as initial pose.}
    \begin{tabularx}{0.9\linewidth}{l*{6}{|>{\centering\arraybackslash}X >{\centering\arraybackslash}X}}
    \hline
        &
        \multicolumn{2}{c|}{PVN3D\cite{he2020pvn3d}}&
        \multicolumn{2}{c|}{+GeoTransformer\cite{qin2022geometric}}&
        \multicolumn{2}{c|}{+ICP-plane\cite{chen1992object}}& 
        \multicolumn{2}{c|}{+GICP\cite{segal2009generalized}}&
        \multicolumn{2}{c|}{+Colored 6D \cite{johnson1999registration}}&
        \multicolumn{2}{c}{+Ours}\\
        \hline
        ADD(S)-N & 0.1 & 0.05 & 0.1 & 0.05 & 0.1 & 0.05 & 0.1 & 0.05 & 0.1 & 0.05 & 0.1 & 0.05\\
        \hline
        ape & 57.5  & 15.7  & \underline{57.2} & 15.8 & 62.1  & 18.1  & 62.3  & 19.6 & 62.5  & 19.1  & \textbf{63.8}  & \textbf{22.1} \\ 
        can & 91.8  & 40.9  & 92.3  & 64.5 & 92.5  & 60.6  & 92.4  & 60.5 & 92.8  & 61.7  & \textbf{94.8}  & \textbf{65.7} \\ 
        cat & 32.4  & 5.2  & 37.2  & 17.4 & 37.6  & 14.2  & 37.2  & 14.2 & 37.4  & 14.2  & \textbf{39.8}  & \textbf{14.7} \\ 
        driller & 76.6  & 19.0  & 80.1  & \textbf{66.4} & 77.5  & 63.6  & \underline{75.0}  & 64.2 & 81.5  & 65.3  & \textbf{82.3}  & 57.8 \\ 
        duck & 33.9  & 2.3  & 53.3  & \textbf{16.9} & 50.9  & 12.8  & 54.9  & \textbf{15.2} & 50.9  & 13.1  & 50.5  & 13.3 \\ 
        \textbf{eggbox} & \textbf{64.6}  & 15.0  & \underline{58.6}  & 19.9 & \underline{60.9}  & 19.3 & \underline{62.0}  & 20.8 & \underline{61.2}  & 19.2  & \underline{63.7}  & \textbf{21.2} \\ 
        \textbf{glue} & 69.8  & 43.3  & \underline{67.8}  & 46.4 & \underline{69.4}  & 50.1 & \underline{67.7}  & 51.2 & \underline{67.9}  & 46.1  & \textbf{70.8}  & \textbf{53.9} \\ 
        holepuncher & 70.1  & 18.8  & 79.5  & 48.6 & 79.8  & 38.7  & 81.9  & 48.4 & 80.5  & 44.2  & \textbf{83.7}  & \textbf{49.1} \\ \hline
        Average & 62.1  & 20.0  & 65.8  & 37.0  & 66.3  & 34.7  & 66.7  & 36.8  & 66.8  & 35.4  & \textbf{68.7}  & \textbf{37.2} \\ 
        \hline
    \end{tabularx}
    \label{tab:occ_comparison2}
\end{table*}

\color{black}

\color{black}
Table \ref{tab:occ_comparison1} and Table \ref{tab:occ_comparison2} show the experimental results on the Occlusion LINEMOD dataset using FFB6D and PVN3D outputs as initial poses, respectively. The results demonstrate that when using FFB6D as the initial pose, our method outperforms GICP and point-to-plane ICP by 0.9\% in ADD(S)-0.1 and improves the initial results by 2.0\%. Moreover, our method also outperforms GICP by 0.7\% in ADD(S)-0.05 and improves the initial results by 6.9\%. In contrast, all other ICP variants perform worse than the initial results for almost half of the objects in both ADD(S)-0.1 and ADD(S)-0.05, although their overall mean is higher than the initial results. These results indicate that our proposed method can maintain stability under severe occlusion and significantly improve the overall performance compared to other ICP methods. GeoTransformer performs better on ADD(S)-0.05 than on ADD(S)-0.1, achieving optimal results on two object categories. In summary, its performance has improved compared to the YCB-Video dataset. However, there is still a gap between its performance and our method, and the improvement is not consistent enough.
\color{black}

\begin{figure}[]
    \centering
    \includegraphics[width=\linewidth]{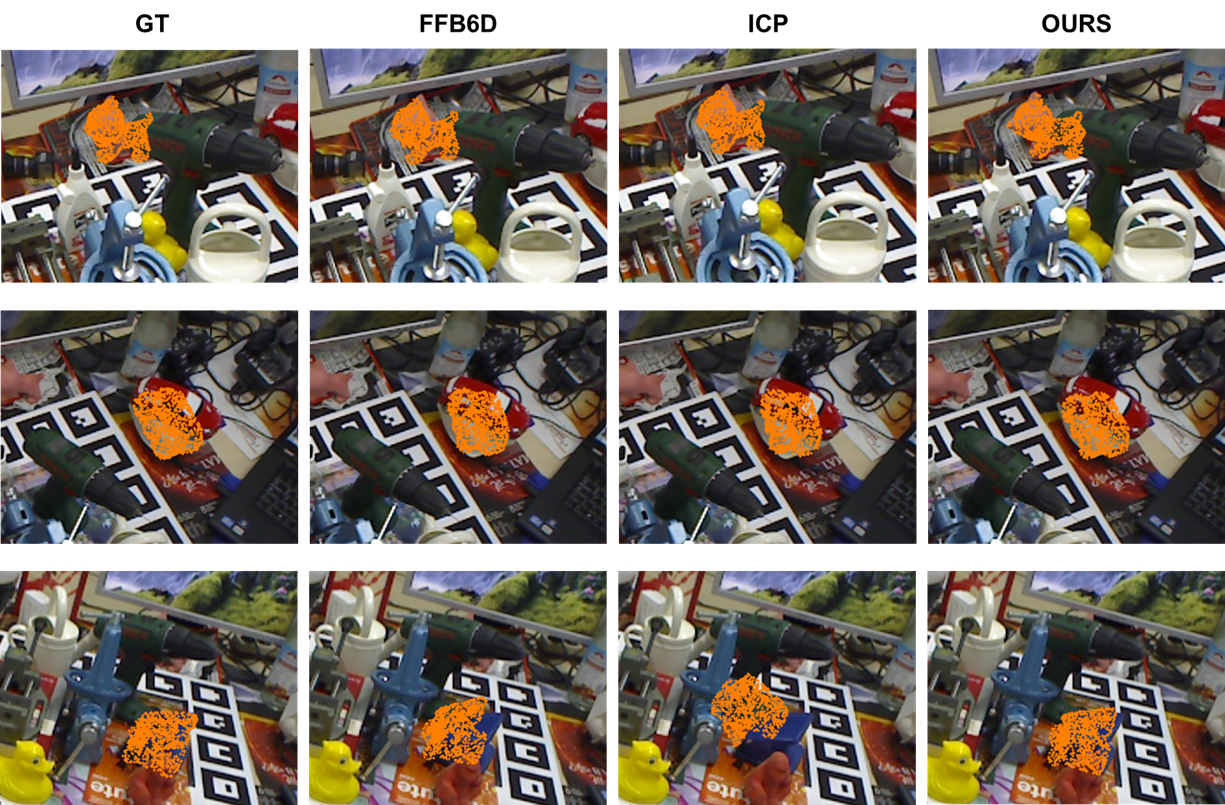}
    \caption{Qualitative results on the Occlusion LineMOD Dataset. Ours can outperform other methods with both high-precision initial pose (row 1) and low-precision initial pose (rows 2 and 3) with higher accuracy.}
    \label{fig:vis2}
\end{figure}

\begin{figure}
    \centering
    \includegraphics[width=1.0\linewidth]{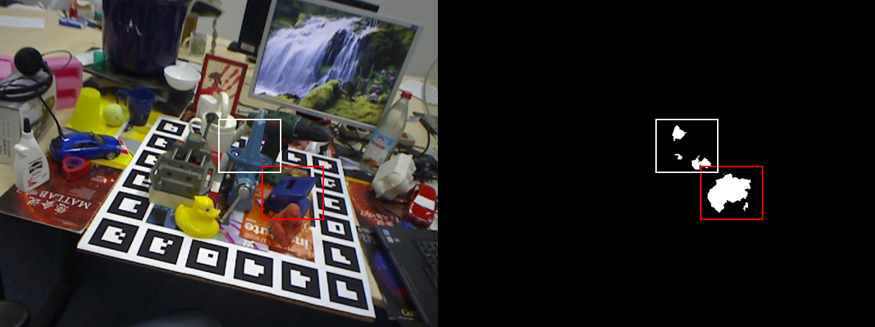}
    \caption{Incorrect segmentation results lead to poor performance of the ICP method. Valid masks are marked by a red box, and incorrect masks by a white box.}
    \label{fig:mask}
\end{figure}

Figure \ref{fig:vis2} shows the qualitative results on the Occlusion LINEMOD dataset, with FFB6D output as the initial pose. Our proposed method significantly outperforms the ICP method in severely occluded scenes, which can be attributed to the completed point cloud added by our method and the step-by-step keypoint optimization strategy. In addition, it is clear that the third row of the ICP method not only failed to effectively optimize the initial pose but produced even worse results. This can be attributed to the severe occlusion and cluttered scenes, which cause a significant deviation between the given segmentation result and the ground truth label, thereby disrupting the post-processing optimization. Figure \ref{fig:mask} shows the given segmentation result in this scene, where valid masks are marked by a red box, and incorrect masks by a white box. It is evident that the inaccurate labeling in the white box has a significant impact on the performance of the ICP method, resulting in erroneous optimization outcomes. Conversely, our method can effectively mitigate the impact of inaccurate segmentation by leveraging completed point clouds and local registration strategies.

In the experiment using PVN3D output as the initial pose, our method and the ICP variants showed a greater improvement in the initial results compared to the previous set of experiments. In ADD(S)-0.1, our method outperformed the Colored 6D ICP by 1.9\%, achieving an improvement of 6.6\% over the initial result. In ADD(S)-0.05, our method outperformed GICP by 0.4\%, achieving an improvement of 17.2\% over the initial result. Notably, in the stricter ADD(S)-0.05, the effect of post-processing optimization is greatly improved, which highlights the importance of further optimizing the results at the end. In this experiment, our method achieved the best results for most objects, with the eggbox being the only exception where it performed lower than the initial result on ADD(S)-0.1. Although the point-to-point ICP method outperformed our method by 7.6\% on the ADD(S)-0.05 of the driller, our method showed superior performance and stability over the ICP variants in most of the other results.

It is worth noting that while the PVN3D's pose estimation results may not be as accurate as those of FFB6D, the post-processing optimization still produced lower results for ADD(S)-0.1 compared to FFB6D, whether using ICP variants or our method. However, the optimization results for FFB6D are still higher than those of PVN3D when evaluating using a stricter metric like ADD(S)-0.05. This may be due to the differences in segmentation results obtained by different methods, which can result in variations in the quality of the point cloud used in registration, ultimately affecting the overall process. Nevertheless, the estimated results for FFB6D are still significantly higher than those of PVN3D after optimization when evaluating using a stricter metric, which is consistent with expectations.

\color{black}
\subsection{RGB Mask Experiment}

\begin{table}[t]
    \centering
    \color{black}
    \caption{RGB Mask Experiment}
    \begin{tabular}{c|cccc}
        \hline
         \diagbox[dir=NW]{Method}{Mask ratio} & 25\%  &  50\% & 75\% & 100\% \\ \hline
        PVN3D\cite{he2020pvn3d}+Ours+YCB & 92.5 & 92.4 & 92.4 & 92.1 \\ \hline
        FFB6D\cite{he2021ffb6d}+Ours+YCB & 93.4 & 93.4 & 93.3 & 93.1 \\ \hline
        PVN3D\cite{he2020pvn3d}+Ours+LO & 68.5 & 68.2 & 68.0 & 67.1 \\ \hline
        FFB6D\cite{he2021ffb6d}+Ours+LO & 66.7 & 66.7 & 66.3 & 65.9 \\ \hline
    \end{tabular}
    \label{tab:rgbmask}
\end{table}

To investigate the impact of texture information on our proposed method, we randomly masked a proportion of RGB information from points in the synthetic data from the YCB-Video and Linemod Occlusion datasets, gradually increasing the masking ratio until all RGB information was obscured.

In Table \ref{tab:rgbmask}, LO denotes the Linemod Occlusion dataset, while YCB represents the YCB-Video dataset. The experimental results demonstrate several points. Firstly, our approach can maintain a stable improvement even under conditions of low texture and no texture. Secondly, the effectiveness of our method experiences the largest decrease when transitioning from low texture (75\%) to no texture (100\%). Finally, in the absence of texture, only the keypoint module in our approach remains functional, and experimental results show that this module still provides a stable improvement.

\subsection{Experiments on symmetric objects}

\begin{table}[h]
    \centering
    \color{black}
    \caption{Comparison of symmetric objects}
    \begin{tabular}{c|c}
        \hline
        & symmetric objects dataset \\ \hline
        PVN3D\cite{he2020pvn3d} & 84.1  \\ 
        PVN3D\cite{he2020pvn3d}+ICP-point\cite{besl1992method} & 83.8  \\ 
        PVN3D\cite{he2020pvn3d}+ICP-plane\cite{chen1992object} & 85.4  \\ 
        PVN3D\cite{he2020pvn3d}+GICP\cite{segal2009generalized} & 84.8  \\ 
        PVN3D\cite{he2020pvn3d}+Colored 6D\cite{johnson1999registration} & 84.0  \\ 
        PVN3D\cite{he2020pvn3d}+Ours & 85.4  \\ 
        FFB6D\cite{he2021ffb6d} & 86.5  \\ 
        FFB6D\cite{he2021ffb6d}+ICP-point\cite{besl1992method} & 85.9  \\ 
        FFB6D\cite{he2021ffb6d}+ICP-plane\cite{chen1992object} & 85.9  \\ 
        FFB6D\cite{he2021ffb6d}+GICP\cite{segal2009generalized} & 85.7  \\ 
        FFB6D\cite{he2021ffb6d}+Colored 6D\cite{johnson1999registration} & 85.9  \\ 
        FFB6D\cite{he2021ffb6d}+Ours & \textbf{86.8}  \\ 
        \hline
    \end{tabular}
    \label{tab:cd}
\end{table}

We conducted a comparative experiment by selecting seven symmetrical objects from the YCB-Video dataset and LineMOD datasets to compose a symmetrical object dataset. This dataset comprises the bowl, wood block, large clamp, extra large clamp, and foam brick from the YCB dataset, as well as the egg box and glue from the LineMOD dataset. The experimental results indicate that, in the case of symmetrical objects, the performance of refinement methods is generally unsatisfactory. Most refinement approaches even deteriorate the accuracy of FFB6D. Only our proposed method consistently maintains a significant and stable improvement, further corroborating the robustness of our approach.

\color{black}

\subsection{Experiments about point cloud completion}

\begin{table*}[h]
    \centering
    \caption{Experimental results using different point cloud completion methods on the YCB-Video dataset}
    \begin{tabular}{c| c| c c c| c c c| c c c}
        \hline
        & Init &
        \multicolumn{3}{c|}{ICP-point \cite{besl1992method}}&
        \multicolumn{3}{c|}{Colored 6D ICP \cite{johnson1999registration}}& 
        \multicolumn{3}{c}{CIKP}\\ \hline
        \diagbox[dir=NW]{Init Pose}{Method} & / & None & PCN & Ours & None & PCN & Ours & None & PCN & Ours \\ 
        \hline
        FFB6D & 92.9 & 92.8 & 92.7 & \textbf{92.9} & 92.8 & 92.9 & \textbf{93.1} & 93.2 & 92.9 & \textbf{93.4} \\ 
        PVN3D & 91.5 & 91.9 & 92.0 & \textbf{92.2} & 92.0 & 92.0 & \textbf{92.2} & 92.2 & 91.8 & \textbf{92.5} \\ 
        \hline
    \end{tabular}
    \label{tab:pcld}
\end{table*}

In addition to comparing the entire PCKRF framework, we also conduct experiments on the point cloud completion method. For our pipeline, we are particularly concerned with the details around keypoints, so we designed a keypoint detector as a decoder to enhance the details around keypoints. However, we also found that our detector did not perform well for overly complex completion frameworks, and even had a counterproductive effect. Based on this, we chose PCN as our backbone. 

\color{black}
Table \ref{tab:seedpcn} is performed on the YCB dataset. We employed SeedFormer \cite{zhou2022seedformer}, a Transformer-based point cloud completion method, along with the traditional PCN \cite{yuan2018pcn} as the backbone while keeping other modules unchanged apart from the completion network. Experimental results demonstrated that the inclusion of SeedFormer for point cloud completion had minimal impact on the final outcome. Surprisingly, the direct application of PCN even showed adverse effects on the results. However, when combined with our custom-designed completion network, it positively influenced the accuracy. We postulate that this is primarily due to PCN's superior adaptability compared to Transformer in handling bidirectional fusion of RGB and point cloud information for the task of point cloud completion.

\begin{table}[h]
    \centering
    \color{black}
    \caption{Comparison of different point cloud completion methods}
    \begin{tabular}{c|cc}
        \hline
        & \multicolumn{2}{c}{YCB-Video Dataset} \\ \cline{2-3}
        & FFB6D & PVN3D \\ \hline
        None & 93.2 & 92.2 \\ 
        PCN \cite{yuan2018pcn} & 92.9 & 91.8 \\ 
        SeedFormer \cite{zhou2022seedformer} & 93.2 & 92.0 \\ 
        SeedFormer \cite{zhou2022seedformer}+Ours & 93.2 & 92.0 \\ 
        PCN+Ours & \textbf{93.4} & \textbf{92.5} \\ 
        \hline
    \end{tabular}
    \label{tab:seedpcn}
\end{table}

\color{black}

Table \ref{tab:pcld} shows the experimental results of applying point cloud completion methods to point cloud registration on the YCB-Video dataset, with ADD(S) AUC serving as the evaluation metric. We evaluate the impact of three methods: not using any point cloud completion (None), using the PCN method, and using our proposed method (Ours) with the point-to-point ICP, Colored 6D ICP, and CIKP methods, respectively. The results show that regardless of whether the initial pose estimation results of FFB6D or PVN3D are used, using the PCN method as the point cloud completion method is similar to the optimization results without using point cloud completion for point-to-point ICP and Colored 6D ICP methods. However, for the CIKP method, the result is the opposite and PCN gets lower performance than without completion. On the contrary, after applying our proposed point cloud completion method, the results of the three methods are improved to varying degrees compared to those without a point cloud. In particular, the CIKP method registers the point cloud near the keypoint. Without completion, only visible points near the keypoints participate in optimization. With completion, all keypoints (visible or not), participate in optimization. As a result, the impact of completed point clouds on the CIKP method is more significant than on the ICP and Colored 6D ICP methods. Therefore, the effect of the CIKP method drops significantly when the PCN method is adopted, while the other methods are not affected as much.

The scale of the sampled point cloud has a significant impact on point cloud registration. To determine the appropriate scale for participating in registration, we utilized the PVN3D output as the initial pose on the YCB-Video dataset. The influence of different point cloud scales on the point-to-point ICP method and CIKP method is shown in Table \ref{tab:pcld_sz}. It can be observed that both the point-to-point ICP method and the CIKP method achieved the best results when the point cloud size was set to 3000. As a result, we selected a downsampling threshold $m_2 = 3000$ in Section \ref{sub:details} to ensure consistency with the best-performing approaches.


\subsection{Ablation Study}

In this part, we conduct ablation experiments on the point cloud completion network and the CIKP method respectively.

\begin{table}[h]
    \centering
    \caption{Influence of different point cloud scales on ICP-point and CIKP method}
    \begin{tabular}{ccc}
        \hline
        Size of Point Cloud & ICP-point & CIKP \\ 
        \hline
        2000 & 91.94 & 91.92\\ 
        3000 & \textbf{92.19} & \textbf{92.47} \\ 
        5000 & 92.16 & 92.22 \\ 
        \hline
    \end{tabular}
    \label{tab:pcld_sz}
\end{table}

\paragraph{Completion Network} Table \ref{tab:ablation_network} shows the ablation study of the point cloud completion network on the YCB-Video dataset. The table includes three modules: FFB, DF, and KPDEC, which respectively stand for Full Flow Bidirectional fusion module, DenseFusion feature fusion module, and the keypoint detection module used only during the training process. The first row represents the original PCN network without any modification. The results show that all three modules effectively improve the performance of the network. The FFB module allows for the full fusion of texture and point cloud features of the object in each pixel, while the feature is extended by stacking DF modules to prevent the loss of critical information when concatenating with global features of the point cloud. The KPDEC module, as previously shown, leads to a more accurate pose estimation when the overall quality of the completed point cloud is not much different.

\begin{table}
    \centering
    \caption{Ablation study for Completion Network on the YCB-Video dataset. KPDEC means keypoint detector block.}
    \begin{tabular}{c|c|c|c}
    \hline
        FFB & DF & KPDEC & ADD(S) \\
        \hline
        & & & 92.7 \\
        \checkmark & & &93.1 \\
        \checkmark & & \checkmark & 93.2 \\
        \checkmark & \checkmark & \checkmark & \textbf{93.4} \\
        \hline
    \end{tabular}
    \label{tab:ablation_network}
\end{table}

\begin{table}
    \centering
    \caption{Ablation study for CIKP on the YCB-Video dataset. KP:refinement by keypoints. Pcld:using completion network. }
    \begin{tabular}{c|c|c|c|c|c|c} 
    \hline
         & Init Pose & KP & Color & Pcld & Rotation & ADD(S)  \\
         \hline
        \multirow{6}{*}{FFB6D} & \multirow{6}{*}{92.9} & & \checkmark & & & 93.0 \\
        & & \checkmark & & & & 93.1 \\
        & & \checkmark & \checkmark & & & 93.2 \\
        & & \checkmark & & \checkmark& & 93.2 \\
        & & \checkmark& \checkmark& \checkmark& & \textbf{93.4} \\
        & & \checkmark& \checkmark& \checkmark& \checkmark & 91.2 \\
        \hline
        \multirow{4}{*}{PVN3D} & \multirow{4}{*}{91.5} & \checkmark & & &  &92.1 \\
        & & \checkmark & \checkmark & & & 92.2 \\
        & & \checkmark & & \checkmark& & 92.4 \\
        & & \checkmark& \checkmark& \checkmark& & \textbf{92.5} \\
        \hline
    \end{tabular}
    \label{tab:ablation_cikp}
\end{table}

\paragraph{CIKP} Table \ref{tab:ablation_cikp} shows the ablation study of the CIKP approach with various initial poses on the YCB-Video dataset. The abbreviation KP stands for the iterative keypoint optimization method for pose estimation; while in its absence, the entire point cloud is optimized iteratively, similar to ICP. Color stands for the use of color information, while Pcld stands for the utilization of the completed point cloud generated by our completion network. Rotation refers to the consideration of both translation and rotation terms during keypoint optimization. The results indicate that when considering the rotation item, the ADD(S) of the CIKP method decreases significantly, suggesting that overfitting of the optimal transformation of the local point cloud around the keypoint is a significant issue. This issue is resolved by solely considering the optimal translation transformation during each keypoint’s pose optimization. Regarding other ablation items, it is evident that when using the predicted pose of FFB6D as the initial pose, all modules contribute similarly to the overall result, and the iterative optimization of each keypoint ensures optimization process stability. The use of color information can introduce texture information to solve the optimization problem of regular-shaped but rich-textured objects, whereas the completed point cloud provides occluded point cloud information that is absent in the input information. In addition, the results obtained by using the predicted pose of PVN3D as the initial pose show that the performance of adding completed point clouds is superior to that of adding color information. This is due to unsatisfactory cloud segmentation results, which affect the performance of all methods, including the post-processing optimization. However, this issue is alleviated after incorporating the completed point cloud.

\subsection{Limitations}


\begin{table}[t]
    \centering
    \color{black}
    \caption{Comparison of pose refinement methods with different initial pose accuracy on YCB-Video dataset}
    \begin{tabular}{ccccc}
        \hline
        Method & ICP-point &  ICP-plane & GICP & Ours \\ \hline
        Level1 & 92.2 & 91.5 & 91.9 & \textbf{92.3} \\ \hline
        Level2 & 90.3 & 89.7 & 88.9 & 88.6 \\ \hline
        Level3 & 87.4 & 85.6 & 84.3 & 72.1 \\ \hline
    \end{tabular}
    \label{tab:dipa}
\end{table}

\begin{table}[t]
    \centering
    \caption{Time Comparison of pose refinement methods}
    \begin{tabular}{ccccc}
        \hline
        Method & ICP-point & Ours & GICP & ICP-plane \\ \hline
        Time(s/object) & 0.77 & 0.66 & 0.53 & \textbf{0.48} \\ \hline
    \end{tabular}
    \label{tab:time}
\end{table}

\color{black}
We conduct a comparison of pose refinement methods with different initial pose accuracy on the ycb-video dataset as shown in Table \ref{tab:dipa}. Level1,2,3 respectively represent a fixed initial pose difference of 5 degrees and 10 centimeters, 10 degrees and 20 centimeters, and 15 degrees and 30 centimeters. Experimental results in Table \ref{tab:dipa} reveal that as the initial pose accuracy diminishes, the effectiveness of our method rapidly declines. Conversely, methods like ICP exhibit better performance than our approach under these conditions. The reason is that when our method selects the point cloud around the keypoint, it chooses from the vicinity of the keypoint previously selected in the object coordinate. If the initial pose is too far from the ground truth, our method will not be able to select enough points for refinement and effective optimization. Based on this limitation, we also speculated whether point cloud noise would have a more significant impact on our method. However, during testing, it was found that as noise increased, the effectiveness of both our method and the registration method declined at a similar rate, showing no significant difference. In future work, we may introduce the superpoint method \cite{zhou2021patch2pix,qin2022geometric} to dynamically select the point cloud for the refinement process from object models.
\color{black}


Table \ref{tab:time} shows the time comparison of ICP variants and our method. It can be observed that our method performs significantly more work than the ICP method in a single iteration, but the overall time is slightly faster than the ICP-point method, indicating that our method can converge relatively quickly. Hence, our method can substitute the original ICP method when it is necessary to optimize high-precision pose estimation. However, our method is still slower than the GICP and point-to-plane method and is not suitable enough for real-time environments, which is another direction that can be improved in the future.

\section{Conclusions}
\color{black}
In this paper, we proposed a pose refinement pipeline PCKRF that integrates the point cloud completion network and CIKP method. The point cloud completion network incorporates a keypoint detection module during the training process to enhance the sensitivity of the completed point cloud, thereby improving the performance of pose refinement. The CIKP method employs a keypoint refinement strategy and incorporates color information to enhance the accuracy and stability of the refinement results. Experiments show that all novel components are effective, and our method outperforms existing refinement methods in optimizing high-precision pose estimation methods on both the YCB-Video dataset and the Occlusion LineMOD dataset. Notably, the results unequivocally demonstrate that our method can seamlessly integrate with the majority of existing pose estimation techniques, resulting in significantly enhanced performance across most cases. Additionally, our approach consistently yields promising outcomes, even in challenging situations characterized by textureless and symmetrical objects. Our experiments also demonstrate that current learning-based point cloud registration methods are not suitable enough for pose refinement. In future work, we will explore the possibility of applying learning-based registration approaches to pose refinement.
\color{black}

\addtolength{\textheight}{-0.0cm}   







\bibliographystyle{IEEEtran}
\bibliography{arxiv}

\begin{IEEEbiography}[{\includegraphics[width=1in,height=1.25in,clip,keepaspectratio]{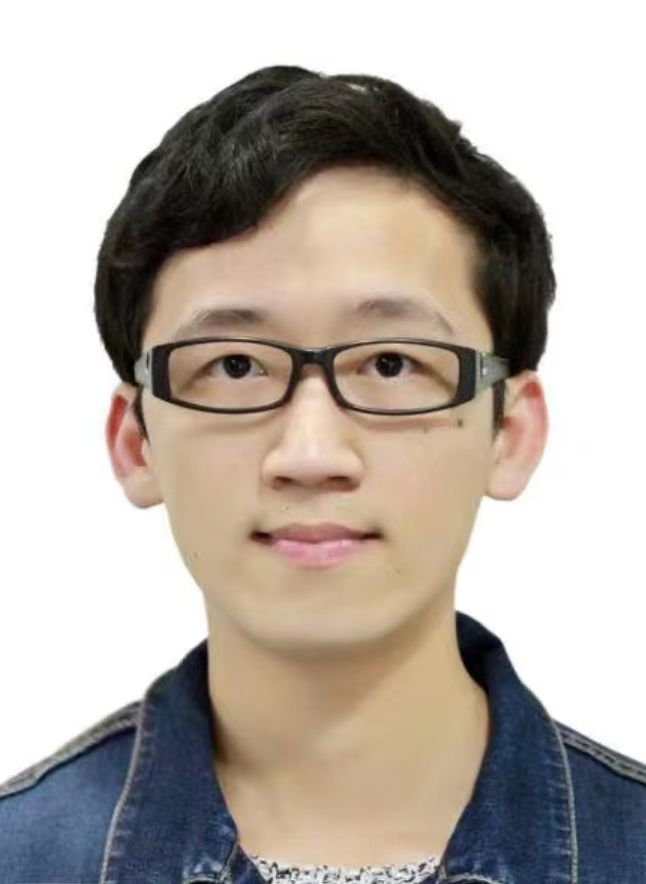}}]{Yiheng Han} is currently an assistant professor with the Faculty of Information Technology, Beijing University of Technology, Beijing, China. He received his B.Eng. degree from Jilin University, China, in 2018, and his Ph.D. degrees from Tsinghua University, China, in 2023. His research interests include robot active vision, motion planning and computer vision.
 \end{IEEEbiography}

\begin{IEEEbiography}[{\includegraphics[width=1in,height=1.25in,clip,keepaspectratio]{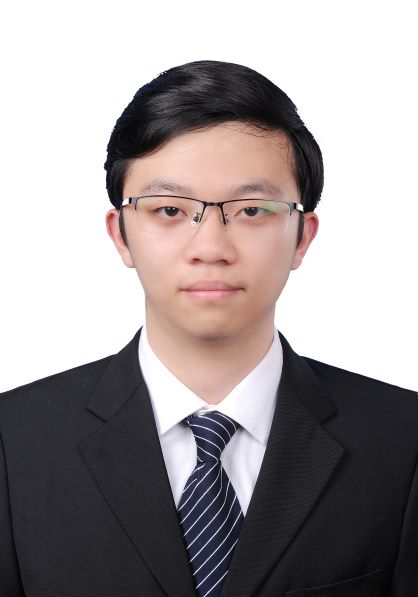}}]{Irvin Haozhe Zhan} is a master degree candidate  with the Department of Computer Science and Technology, Tsinghua University, China. He received his B. Eng. degree from Tsinghua University in 2020. His research interests include computer vision, robotics and deep learning.
 \end{IEEEbiography}

\begin{IEEEbiography}[{\includegraphics[width=1in,height=1.25in,clip,keepaspectratio]{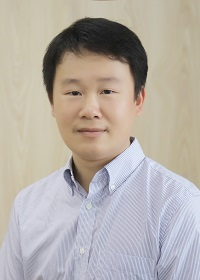}}]{Long Zeng} is an associate professor with the Shenzhen International Graduate School, Tsinghua University, China. He received his M.Phil degree from Zhejiang Unversity, China, in 2007, and the PhD degree from the Hong Kong University of Science and Technology, Hong Kong, China, in 2012. His research interests include intelligent manufacturing, computer-aided design, and robots. For more information, visit https://www.sigs.tsinghua.edu.cn/cl/main.htm
 \end{IEEEbiography}

\begin{IEEEbiography}[{\includegraphics[width=1in,height=1.25in,clip,keepaspectratio]{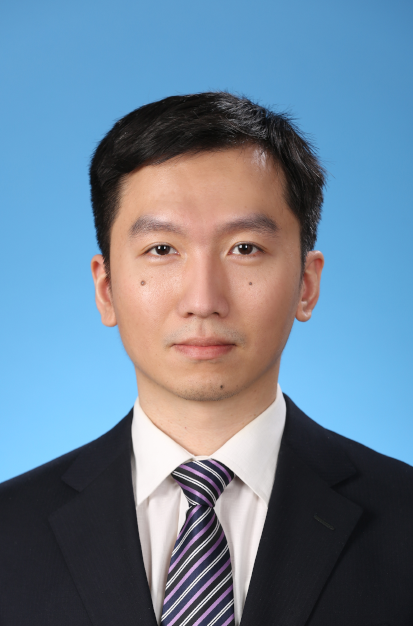}}]{Yu-Ping Wang} is an associate professor at the School of Computer Science and Technology, Beijing Institute of Technology. He received his B.S. degree from Northeastern University, China, in 2002, and his M.E. and Ph.D. degrees from Tsinghua University, China, in 2005 and 2009, respectively. His research interests include robot operating system, and distributed robotic system.
 \end{IEEEbiography}

\begin{IEEEbiography}[{\includegraphics[width=1in,height=1.25in,clip,keepaspectratio]{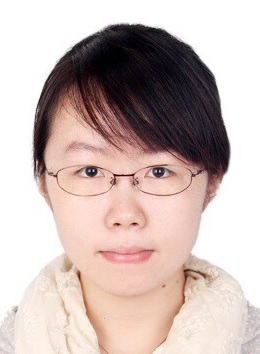}}]{Ran Yi} is an assistant professor with the Department of Computer Science and Engineering, Shanghai Jiao Tong University. She received the BEng degree and the PhD degree from Tsinghua University, China, in 2016 and 2021. Her research interests include computer vision, computer graphics and computational geometry.
 \end{IEEEbiography}

\begin{IEEEbiography}[{\includegraphics[width=1in,height=1.25in,clip,keepaspectratio]{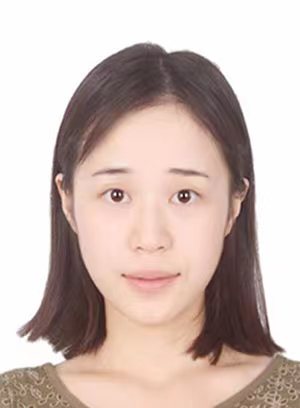}}]{Minjing Yu} received the BE degree from Wuhan University, Wuhan, China, in 2014, and the PhD degree from Tsinghua University, Beijing, China, in 2019. She is currently an associate professor with the College of Intelligence and Computing, Tianjin University, China. Her research interests include computer graphics, artificial intelligence, and cognitive computation.
 \end{IEEEbiography}

\begin{IEEEbiography}[{\includegraphics[width=1in,height=1.25in,clip,keepaspectratio]{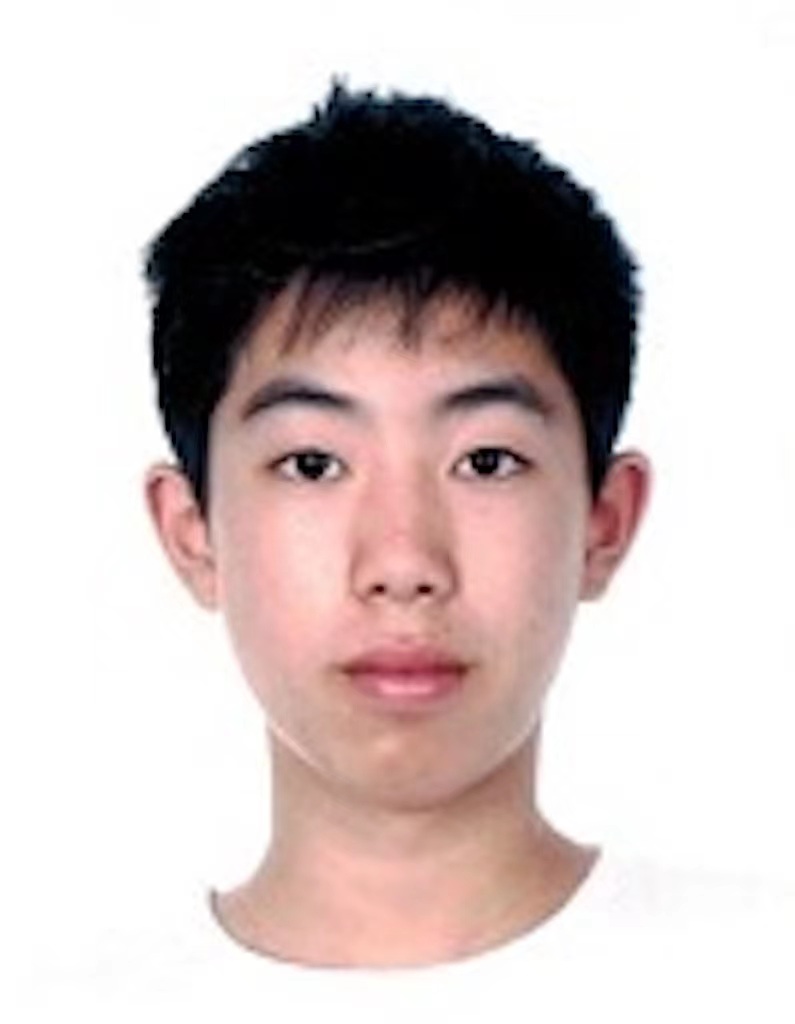}}]{Matthieu Gaetan Lin} is PhD candidate with the Department of Computer Science and Technology, Tsinghua University, China. His research interests include computer vision, intelligent media processing, and human-computer interaction.
 \end{IEEEbiography}

\begin{IEEEbiography}[{\includegraphics[width=1in,height=1.25in,clip,keepaspectratio]{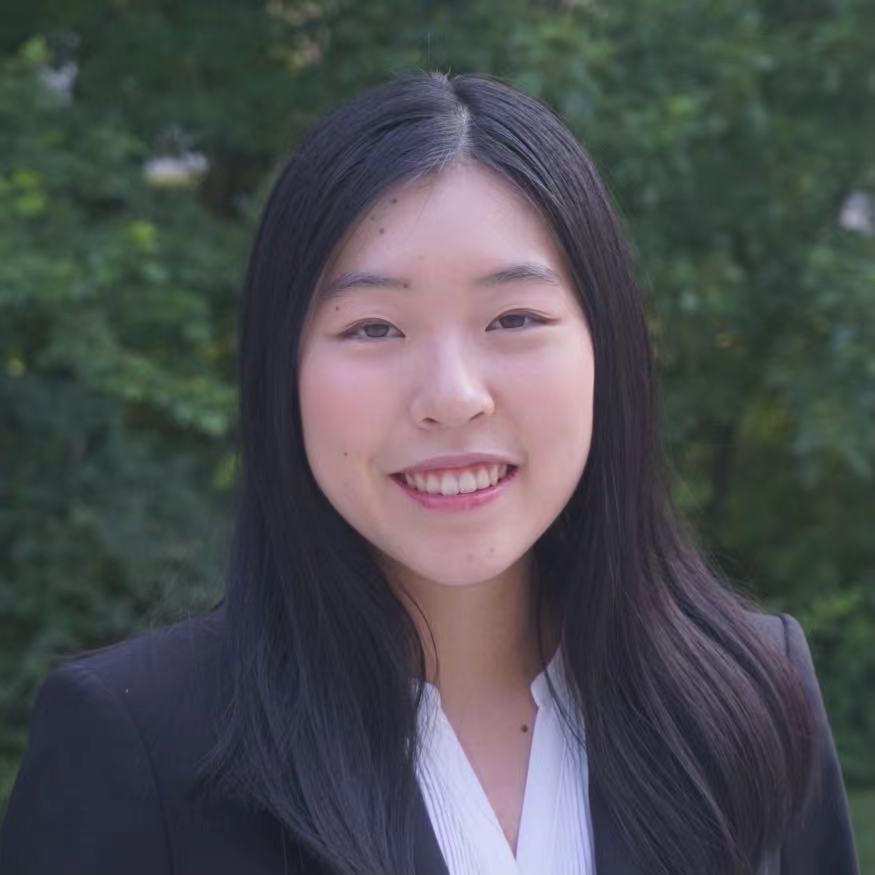}}]{Jenny Sheng} is a master degree candidate with the Department of Computer Science and Technology, Tsinghua University, China. She received her B.S.E. degree in Computer Science from Princeton University in 2022. Her research interests include computer vision, intelligent media processing, and human-computer interaction.
 \end{IEEEbiography}

\begin{IEEEbiography}[{\includegraphics[width=1in,height=1.5in,clip,keepaspectratio]{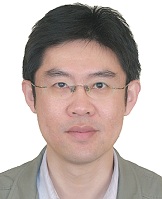}}]{Yong-Jin Liu} is a professor with the Department of Computer Science and Technology, Tsinghua University, China. He received the BEng degree from Tianjin University, China, in 1998, and the PhD degree from the Hong Kong University of Science and Technology, Hong Kong, China, in 2004. His research interests include computer vision, computer graphics and computer-aided design. For more information, visit \url{http://cg.cs.tsinghua.edu.cn/
people/~Yongjin/Yongjin.htm}.
 \end{IEEEbiography}

\vfill

\end{document}